\def\impactCameraReadyAuthors{%
  Rongze Tang\textsuperscript{\rm 1,\rm 2}\equalcontrib,
  Jianjie Fang\textsuperscript{\rm 3}\equalcontrib,
  Zhaolu Wang\textsuperscript{\rm 3},
  Ziyou Wang\textsuperscript{\rm 3},
  Xvyuan Liu\textsuperscript{\rm 3},\\
  Haisheng Su\textsuperscript{\rm 4},
  Xin Zhang\textsuperscript{\rm 4},
  Wei Wu\textsuperscript{\rm 4},
  Chen Gao\textsuperscript{\rm 2,\rm 3}\corresponding,
  Yong Li\textsuperscript{\rm 3},
  Zhibo Chen\textsuperscript{\rm 1,\rm 2}\corresponding
}
\def\impactCameraReadyAffiliations{%
  \textsuperscript{\rm 1}University of Science and Technology of China
  \quad \textsuperscript{\rm 2}Zhongguancun Academy \\
  \textsuperscript{\rm 3}Tsinghua University
  \quad \textsuperscript{\rm 4}Manifold AI \\
  \makebox[0pt][c]{chgao96@gmail.com, chenzhibo@ustc.edu.cn}%
  \smash{\raisebox{-1.55em}[0pt][0pt]{\makebox[0pt][c]{%
    \impactExternalLink{https://embodiedcity.github.io/IMPACT/}{\textcolor{magenta}{\faGlobe\ \textit{Project Page}}}%
    \hspace{3em}%
    \impactExternalLink{https://github.com/EmbodiedCity/IMPACT.code}{\faGithub\ \textit{Code}}%
  }}}
}

\documentclass[letterpaper]{article}
\usepackage{aaai2027}
\nocopyright
\usepackage[hyphens]{url}
\usepackage{graphicx}
\usepackage{natbib}
\usepackage{caption}
\usepackage{amsmath,amssymb,amsfonts,amsthm}
\usepackage{booktabs}
\usepackage{multirow}
\usepackage{colortbl}
\usepackage{array}
\usepackage{fontawesome5}

\title{IMPACT: Attention Is the Interaction Map for\\Scalable Interaction-Aware World Model Training}

\makeatletter
\newcommand{\impactMultipleCorrespondingAuthors}{\global\aaai@corrmultitrue}
\makeatother
\author{\impactMultipleCorrespondingAuthors\impactCameraReadyAuthors}
\affiliations{\impactCameraReadyAffiliations}

\newcommand{\method}{IMPACT}

\makeatletter
\newcommand{\impact@dest}[1]{\pdfdest name {impact.#1} xyz\relax}
\newcommand{\impact@linkto}[2]{%
  \leavevmode
  \pdfstartlink attr {/Border [0 0 0]} goto name {impact.#1}\relax
  #2%
  \pdfendlink
}
\newcommand{\impactExternalLink}[2]{%
  \leavevmode
  \pdfstartlink attr {/Border [0 0 0]} user {/Subtype /Link /A << /S /URI /URI (#1) >>}\relax
  #2%
  \pdfendlink
}
\renewcommand{\hyper@natanchorstart}[1]{\pdfdest name {impact.cite.#1} xyz\relax}
\renewcommand{\hyper@natanchorend}{}
\renewcommand{\hyper@natlinkstart}[1]{%
  \leavevmode
  \pdfstartlink attr {/Border [0 0 0]} goto name {impact.cite.#1}\relax
}
\renewcommand{\hyper@natlinkend}{\pdfendlink}
\renewcommand{\hyper@natlinkbreak}[2]{%
  \hyper@natlinkend#1\hyper@natlinkstart{#2}%
}
\let\impact@oldlabel\label
\renewcommand{\label}[1]{\impact@dest{#1}\impact@oldlabel{#1}}
\AtBeginDocument{%
  \@ifundefined{ltx@label}{}{%
    \let\impact@oldltxlabel\ltx@label
    \def\ltx@label#1{\impact@dest{#1}\impact@oldltxlabel{#1}}%
  }%
}
\let\impact@plainref\ref
\renewcommand{\ref}[1]{\impact@linkto{#1}{\impact@plainref{#1}}}
\newcommand{\figref}[1]{\impact@linkto{#1}{Figure~\impact@plainref{#1}}}
\newcommand{\tabref}[1]{\impact@linkto{#1}{Table~\impact@plainref{#1}}}
\newcommand{\eqnref}[1]{\impact@linkto{#1}{Equation~\impact@plainref{#1}}}
\newcommand{\eqsref}[1]{\impact@linkto{#1}{Eq.~\impact@plainref{#1}}}
\makeatother

\begin{document}
\maketitle

\begin{abstract}
World models have made remarkable progress in action-conditioned future prediction for embodied agents, yet still struggle to model physically plausible interactions. Existing approaches address this limitation by constraining the generation process with external representations encoding motion, geometry, or semantics. Obtaining these spatiotemporally dense representations typically requires auxiliary estimators or manual annotations, limiting training scalability. We instead revisit the training objective and identify a \textbf{supervision-allocation mismatch} under the globally averaged mean squared error (MSE) denoising objective: prevalent static content dominates the optimization signal, leaving sparse dynamic-object regions critical to interaction generation disproportionately under-supervised. Motivated by this observation, we introduce \textbf{IMPACT}, a scalable \textbf{I}nteraction-aware \textbf{M}odel training framework with \textbf{P}rior-guided \textbf{A}ttention \textbf{C}alibration and \textbf{T}argeting. IMPACT uses cross-attention associated with manipulated-object tokens as an internal spatiotemporal prior for action-conditioned changes. It samples candidate regions from this prior, calibrates them with detached local prediction errors to construct an interaction map, and uses the map to reweight denoising supervision, requiring neither external representations nor inference-time modifications. Extensive experiments on robot-arm and human-hand manipulation, spanning diverse control
modalities and DiT backbones, show that \method{} consistently outperforms the corresponding MSE-trained baselines, improving interaction fidelity, physical plausibility, and visual quality.
\end{abstract}

\begin{figure}[t]
    \centering
    \includegraphics[width=\linewidth]{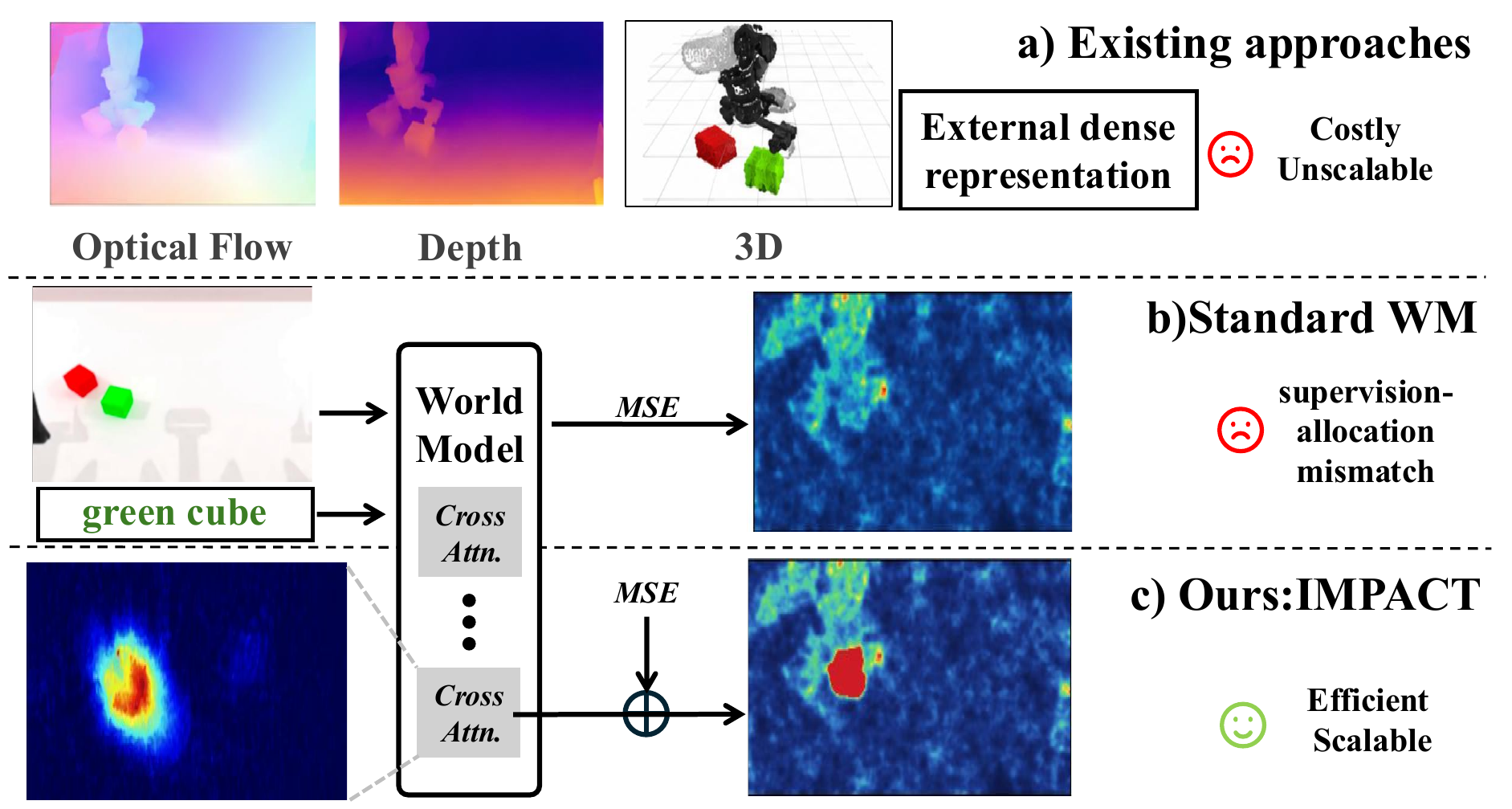}
    \caption{Overview of \method{}.
(a) Existing approaches rely on costly and unscalable external dense representations.
(b) Standard world-model training with uniformly averaged MSE suffers from a supervision-allocation mismatch, under-supervising sparse interaction regions.
(c) \method{} calibrates object-conditioned cross-attention with prediction errors into an interaction map for efficient and scalable interaction-targeted training.}
    \label{fig:overview}
\end{figure}

\section{Introduction}
\label{sec:introduction}

World models have made remarkable progress in predicting future observations
\citep{zhang2026qwenrobotworld,kim2026dwm,fang2026worldscape}
and supporting embodied simulation
\citep{wang2026lifting,gao2025flip}.
Building on large-scale pretrained video generation backbones\citep{yang2024cogvideox,kong2024hunyuanvideo,wan2025wan,agarwal2025cosmos},
they can simulate how an environment evolves under a commanded action, generating action responses in interactive environments
\citep{alayaworld2026,hu2026multiplayer,fang2026iworld}
and robotic manipulation tasks
\citep{zhang2026qwenrobotworld,bi2026motus,zhang2026physisforcing}.
Their imagined futures provide scalable experience for long-horizon planning
\citep{wang2026lifting,gao2025flip},
policy learning
\citep{hu2025vpp,zhen2025tesseract,su2026worldscape},
and policy evaluation
\citep{shang2026worldarena,zhu2025irasim}.
In this role, world models should go beyond visual coherence to simulate physically plausible interactions, yet still suffer from object deformation, discontinuous motion, weak action coupling and inconsistent contact
\citep{zhang2026physisforcing}.

As shown in \figref{fig:overview}a, existing approaches typically address these failures by constraining generation with external representations of interaction dynamics.
Motion-based methods use optical flow or point trajectories to describe object dynamics
\citep{gao2025flip,zhang2026physisforcing} and
geometry-based methods introduce depth, surface normals, reconstructed scenes, or articulated hand meshes
\citep{zhen2025tesseract,kim2026dwm}.
However, obtaining such spatiotemporally dense representations through auxiliary estimators or manual annotations is costly, and the resulting supervision is bounded by the accuracy of these external signals, limiting both training scalability and the achievable interaction quality.

We instead revisit the standard training objective of video world models and identify a \emph{supervision-allocation mismatch}, as illustrated in \figref{fig:overview}b.
Inherited from general video generation, this globally averaged mean squared error (MSE) denoising objective uniformly weights all spatiotemporal positions
\citep{po2025longcontext,zhu2025irasim,kim2026dwm}.
Under uniform weighting, each region contributes to optimization according to its spatial extent rather than its functional importance.
Prevalent static content thus dominates the training signal, leaving sparse dynamic-object regions that carry action-conditioned changes disproportionately under-supervised.
As a result, models may reduce the global denoising loss and generate visually coherent videos while leaving interaction regions under-optimized.

Our key insight is that world models built on large-scale pretrained video generation backbones already contain a spatiotemporal prior for interaction regions.
For manipulation instructions, cross-attention aligns language tokens with spatiotemporal video representations, enabling the attention maps associated with manipulated-object tokens to serve as a spatiotemporal prior for regions likely to undergo action-conditioned changes.
The refined prior provides a natural basis for reweighting denoising supervision, allowing sparse interaction regions to contribute to gradient optimization according to their functional importance rather than their spatial extent, thereby improving interaction generation.
Because this prior is obtained from the model's standard forward pass, it requires no external dense representations, scales readily with training data, and leaves inference unchanged.

Based on this insight, we introduce \textbf{IMPACT}, a scalable \textbf{I}nteraction-aware \textbf{M}odel training framework with \textbf{P}rior-guided \textbf{A}ttention \textbf{C}alibration and \textbf{T}argeting.
As shown in \figref{fig:overview}c, the core idea is to treat object-conditioned cross-attention as an interaction prior, turn it into an interaction map, and use this map to reweight denoising supervision toward interaction regions, thereby mitigating the supervision-allocation mismatch and improving interaction generation.
IMPACT realizes this within a single training step through two complementary components.
In the forward pass, \textbf{Attention Distribution Sampling} (ADS) aggregates the cross-attention of the manipulated-object tokens into a proposal distribution, samples multiple candidate regions, and weights each by its detached local prediction error, calibrating the attention prior with the model's current prediction difficulty to form a precise interaction map.
In the backward pass, \textbf{Interaction-Weighted Supervision} (IWS) uses this map to strengthen denoising supervision on interaction regions while preserving the global objective, and routes gradients so that the interaction-weighted objective updates the non-cross-attention DiT parameters while the attention-producing parameters follow the global objective, preventing the prior from collapsing onto its own signal.

We evaluate IMPACT in two settings: robotic-arm manipulation on WorldArena
\citep{shang2026worldarena} and human-hand manipulation on EgoDex \citep{hoque2025egodex}.
Across both settings, IMPACT surpasses standard uniformly-supervised training (MSE) under the same backbone as well as various baselines, delivering higher interaction fidelity and physical plausibility across diverse embodied scenarios and action conditions.

In summary:
\begin{itemize}
    \item
    We identify a supervision-allocation mismatch in world model training and introduce IMPACT, a scalable training framework that leverages the object-conditioned cross-attention prior to reweight denoising supervision toward interaction regions, without requiring any external dense representations.

    \item
    We realize IMPACT through two complementary designs: ADS evaluates object-conditioned attention proposals using detached local prediction errors and converts these errors into weights to form an interaction map, while IWS uses this map to target denoising supervision, preserve the global objective, and decouple region estimation from regional optimization.

    \item
    We demonstrate across robotic-arm and human-hand manipulation with different control signals and DiT backbones that IMPACT consistently outperforms uniform-MSE training, improving both the physical consistency and visual quality of generated videos.
\end{itemize}

\section{Related Work}
\label{sec:related}

\paragraph{Interactive video generation and world models}
Building on large-scale pretrained video generation backbones, including CogVideoX, HunyuanVideo, Wan, and Cosmos \citep{yang2024cogvideox,kong2024hunyuanvideo,wan2025wan,agarwal2025cosmos}, recent world models inherit rich representations of visual appearance and temporal dynamics, enabling high-fidelity prediction and coherent motion modeling, but these capabilities alone are insufficient to make them useful embodied simulators.
To bridge this gap, recent work introduces diverse control conditions that make world models interactive, including natural-language instructions \citep{xiang2024pandora,zhang2026qwenrobotworld,zhao2025airscape,zhao2026worldvln}, game and camera controls \citep{bruce2024genie,he2025matrixgame2}, robot action trajectories \citep{agarwal2025cosmos,zhu2025irasim}, and articulated hand or body controls \citep{wang2026hand2world,gao2026lomelearninghumanobjectmanipulation}.
However, these methods primarily target overall visual quality and controllable generation, while providing limited constraints on interaction dynamics, resulting in physically implausible behavior within interaction regions.

\paragraph{External representations priors}
To address this limitation, recent methods introduce spatiotemporally dense representations such as optical flow, depth maps, or reconstructed 3D structure to explicitly constrain the generation process. Motion-based approaches exploit optical flow, point trajectories, or latent temporal discrepancies to emphasize dynamic regions \citep{fang2025motionimage,zhang2026physisforcing,wu2026ltd}. Geometry-based methods introduce depth, cross-view 3D structure, or projected robot kinematics as auxiliary targets or structured conditions \citep{tian2026starry,yang2026eawm,liu2025geometryaware}. However, constructing these representations often requires external motion, depth, segmentation, and video-understanding models, sometimes together with manually verified annotations \citep{fang2025motionimage,zhang2026physisforcing,yan2025scar,luo2026cointeract}. These preprocessing costs grow with the size and duration of the training corpus, thereby limiting training scalability.

\section{Method}
\label{sec:method}

\begin{figure*}[t]
    \centering
    \includegraphics[width=\linewidth]{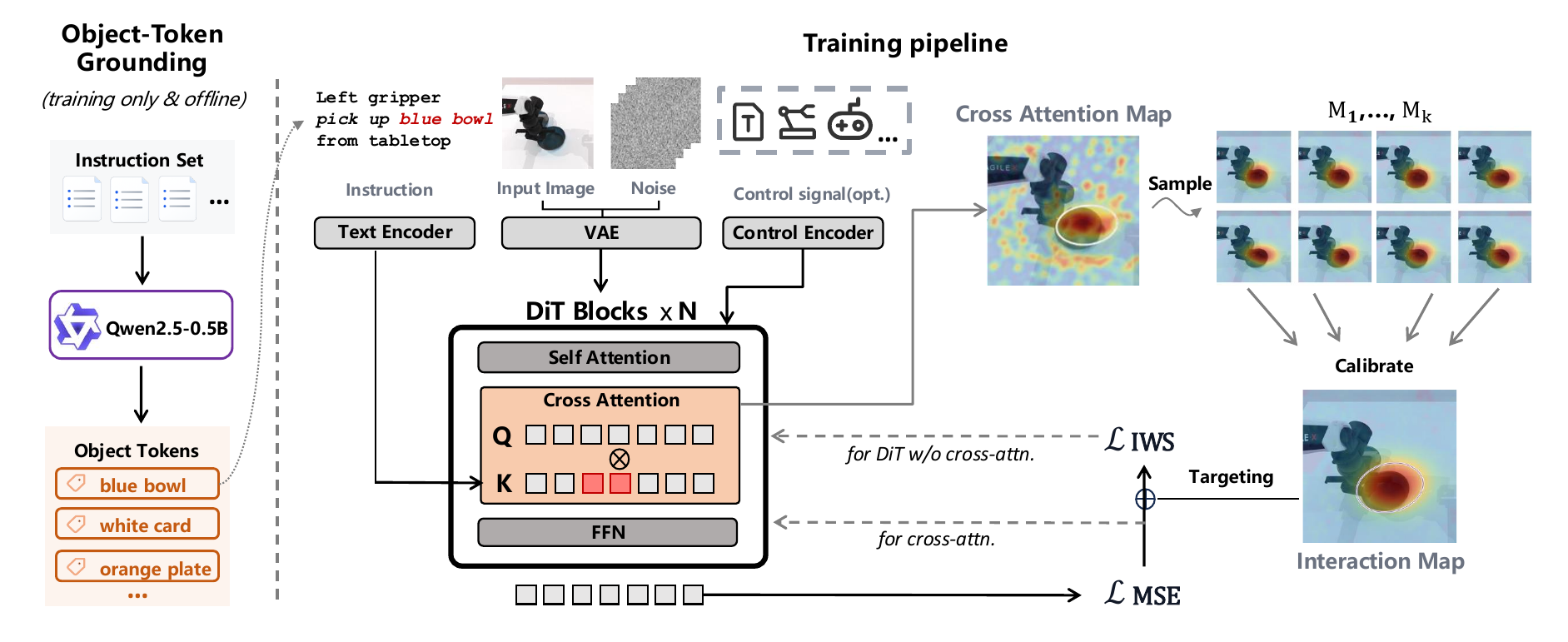}
    \caption{
    Framework of \method{}.
    Object-token grounding (left): a frozen Qwen2.5-0.5B extracts the manipulated-object phrase from the instruction and aligns it to the corresponding object tokens.
    Training pipeline (right): within the DiT blocks, the cross-attention of these object tokens forms a proposal distribution (cross-attention map), from which $\mathbf{M}_1,\dots,\mathbf{M}_8$ candidate regions are sampled and calibrated by their detached local prediction errors into an interaction map.
    The map targets denoising supervision toward interaction regions, and gradient routing optimizes the cross-attention parameters with the global objective $\mathcal{L}_{\mathrm{MSE}}$ and the remaining DiT parameters with the interaction-weighted objective $\mathcal{L}_{\mathrm{IWS}}$.
    }
    \label{fig:framework}
\end{figure*}

\figref{fig:framework} presents an overview of \textbf{\method{}}.
Given a training sample, \method{} identifies the manipulated-object tokens and uses their cross-attention as a spatial prior.
Attention Distribution Sampling(ADS) samples candidate regions from this prior and weights them by detached local prediction errors to construct an interaction map, which Interaction-Weighted Supervision(IWS) uses to target denoising supervision toward interaction-relevant tokens. Gradient routing optimizes the cross-attention parameters with the original global objective and the remaining DiT parameters with the interaction-weighted objective, while all additional operations are training-only and incur no inference-time overhead.

\subsection{Preliminaries}
\label{sec:preliminaries}

We build on a latent video diffusion transformer trained with flow matching \citep{lipman2023flow}.
Given a video $\mathbf{x}_0$, a frozen variational autoencoder maps it into a latent representation
$\mathbf{z}_0 \in \mathbb{R}^{C \times T \times H \times W}$, where $T$, $H$, and $W$ denote the temporal and spatial dimensions.
The conditioning information is denoted by
$\mathbf{c}=(\mathbf{y},\mathbf{x}^{\mathrm{ref}},\mathbf{a})$, where $\mathbf{y}$ is the language instruction, $\mathbf{x}^{\mathrm{ref}}$ is the reference observation, and $\mathbf{a}$ represents the corresponding control signal, such as hand poses or robot trajectories.

For a sampled noise level $\sigma_t \in [0,1]$ and Gaussian noise
$\boldsymbol{\epsilon}\sim\mathcal{N}(\mathbf{0},\mathbf{I})$, the clean latent is interpolated with noise as
\begin{equation}
    \mathbf{z}_t
    =
    (1-\sigma_t)\mathbf{z}_0
    +
    \sigma_t\boldsymbol{\epsilon}.
    \label{eq:flow_interpolation}
\end{equation}
The diffusion transformer $v_{\theta}$ takes
$(\mathbf{z}_t,t,\mathbf{c})$ as input and predicts the velocity target
$\boldsymbol{\epsilon}-\mathbf{z}_0$.
Let
\begin{equation}
    \Omega
    =
    \{1,\ldots,T\}
    \times
    \{1,\ldots,H\}
    \times
    \{1,\ldots,W\}
\end{equation}
denote the set of spatiotemporal latent positions.
The prediction error at position $p\in\Omega$ is
\begin{equation}
    \ell_p
    =
    \frac{1}{C}
    \left\|
    v_{\theta}(\mathbf{z}_t,t,\mathbf{c})_p
    -
    (\boldsymbol{\epsilon}-\mathbf{z}_0)_p
    \right\|_2^2.
    \label{eq:local_error}
\end{equation}
Standard flow-matching training uniformly averages this error over all spatiotemporal positions:
\begin{equation}
    \mathcal{L}_{\mathrm{global}}(\theta)
    =
    \mathbb{E}_{\mathbf{z}_0,\boldsymbol{\epsilon},t}
    \left[
    \frac{1}{|\Omega|}
    \sum_{p\in\Omega}\ell_p
    \right].
    \label{eq:global_objective}
\end{equation}
Consequently, the contribution of a region to optimization is largely determined by its spatial extent.
Small interaction regions therefore receive no additional emphasis despite containing rapid motion, contact transitions, and action-conditioned state changes.
\method{} addresses this supervision-allocation mismatch without changing the underlying flow-matching formulation.

\subsection{Object-Token Grounding}
\label{sec:object_grounding}

The interaction region depends on the object being manipulated.
We therefore begin by grounding the manipulated object in the language instruction.
For an instruction $\mathbf{y}$, a frozen Qwen2.5-0.5B model \citep{qwen2025qwen25} extracts the noun phrase that denotes the manipulated object.
For example, given the instruction ``Right gripper stacks blue bowl on white tabletop,'' the extracted phrase is ``blue bowl.''

Let
\begin{equation}
    \mathbf{e}
    =
    (\mathbf{e}_1,\ldots,\mathbf{e}_N)
\end{equation}
denote the sequence of text embeddings produced by the tokenizer and text encoder.
We align the extracted object phrase with this token sequence and denote its token positions by
\begin{equation}
    \mathcal{O}
    \subseteq
    \{1,\ldots,N\}.
    \label{eq:object_token_set}
\end{equation}
When an object phrase is divided into multiple subword tokens, all matched positions are included in $\mathcal{O}$.
This object-token set provides the semantic anchor used by ADS to extract an object-conditioned spatial distribution.
The grounding stage operates only on the instruction and therefore requires neither visual annotations nor external spatial estimators.

\subsection{Forward: Attention Distribution Sampling}
\label{sec:ads}

ADS converts the semantic object grounding into an interaction map through three operations: object-conditioned attention aggregation, candidate-region sampling, and prediction-error-based weighting.

\paragraph{Object-conditioned attention distribution.}
Within each DiT cross-attention layer, visual queries attend to the text-token sequence.
For block $l$, attention head $h$, visual position $p$, and text-token position $j$, the cross-attention probability is
\begin{equation}
    A_{p,j}^{(l,h)}
    =
    \operatorname{softmax}_{j}
    \left(
    \frac{
    \mathbf{q}_{p}^{(l,h)}
    \mathbf{k}_{j}^{(l,h)\top}
    }{\sqrt{d}}
    \right),
    \label{eq:cross_attention}
\end{equation}
where $d$ is the key dimension.
We sum the probability mass assigned to the object-token set $\mathcal{O}$ and average it over the selected blocks $\mathcal{B}$ and attention heads $\mathcal{H}$:
\begin{equation}
    A_p
    =
    \frac{1}{|\mathcal{B}|\,|\mathcal{H}|}
    \sum_{l\in\mathcal{B}}
    \sum_{h\in\mathcal{H}}
    \sum_{j\in\mathcal{O}}
    A_{p,j}^{(l,h)}.
    \label{eq:object_attention}
\end{equation}
The resulting map
$\mathbf{A}\in[0,1]^{T\times H\times W}$
forms an object-conditioned proposal distribution over the video latent.
Because it is obtained from the standard conditional forward pass, constructing $\mathbf{A}$ requires no additional visual model or spatial annotation.

\paragraph{Candidate-region sampling.}
Instead of using a single deterministic region, ADS samples $K$ spatially coherent candidates around the attention distribution.
We first detach and temper the attention map:
\begin{equation}
    \widetilde{\mathbf{A}}
    =
    \operatorname{clip}
    \left(
    \operatorname{sg}[\mathbf{A}]^{\,\kappa},
    \varepsilon,
    1-\varepsilon
    \right),
    \label{eq:attention_tempering}
\end{equation}
where $\operatorname{sg}[\cdot]$ denotes stop-gradient and
$\kappa$ controls the concentration of the proposal distribution.
We then transform it into logits:
\begin{equation}
    \mathbf{Z}
    =
    \log
    \frac{\widetilde{\mathbf{A}}}
    {1-\widetilde{\mathbf{A}}}.
    \label{eq:attention_logits}
\end{equation}

For each candidate $k\in\{1,\ldots,K\}$, we sample Gaussian noise on a coarse spatiotemporal grid and trilinearly interpolate it to the resolution of $\mathbf{A}$, producing a smooth perturbation field
$\boldsymbol{\eta}_k$.
A soft candidate is generated as
\begin{equation}
    \mathbf{S}_k
    =
    \operatorname{sigmoid}
    \left(
    \frac{
    \mathbf{Z}
    +
    \sigma\boldsymbol{\eta}_k
    }{\tau}
    \right),
    \label{eq:soft_candidate}
\end{equation}
where $\sigma$ controls the perturbation magnitude and $\tau$ is the sampling temperature.
Sampling noise on a coarse grid produces coherent spatial variations rather than independent token-wise perturbations.

To prevent candidate scores from being dominated by differences in region size, every candidate is converted into a fixed-area binary region.
Specifically, we retain the largest $\rho|\Omega|$ values of $\mathbf{S}_k$:
\begin{equation}
    M_{k,p}
    =
    \mathbb{I}
    \left[
    S_{k,p}
    \geq
    \operatorname{TopKThreshold}
    (\mathbf{S}_k,\rho|\Omega|)
    \right],
    \label{eq:fixed_area_candidate}
\end{equation}
yielding
$\mathbf{M}_k\in\{0,1\}^{T\times H\times W}$.
All candidates consequently cover the same number of latent positions and remain directly comparable.

\paragraph{Prediction-error-based weighting.}
ADS evaluates every candidate using the mean local prediction error within that region:
\begin{equation}
    d_k
    =
    \frac{
    \sum_{p\in\Omega}
    M_{k,p}\operatorname{sg}[\ell_p]
    }{
    \sum_{p\in\Omega}M_{k,p}
    +
    \varepsilon
    }.
    \label{eq:candidate_score}
\end{equation}
A higher value of $d_k$ indicates that the candidate covers content that is currently more difficult for the model to predict.
The candidate errors are converted into normalized weights:
\begin{equation}
    \alpha_k
    =
    \frac{
    \exp(\beta d_k)
    }{
    \sum_{j=1}^{K}\exp(\beta d_j)
    },
    \label{eq:candidate_weight}
\end{equation}
where $\beta$ controls the concentration of the weighting distribution.
The candidate regions are then weighted to form the interaction map
\begin{equation}
    \mathbf{G}
    =
    \operatorname{clip}
    \left(
    \sum_{k=1}^{K}
    \alpha_k\mathbf{M}_k,
    0,
    1
    \right).
    \label{eq:interaction_map}
\end{equation}
Through this process, the object-conditioned attention determines where candidate regions are sampled, while the current local prediction error determines their relative contribution to $\mathbf{G}$.
The entire ADS pipeline is computed without gradient tracking, preventing the model from directly modifying the proposal distribution or candidate weights to reduce the regional objective.

In our implementation, we use $K=8$, an attention tempering exponent
$\kappa=0.65$, perturbation scale $\sigma=0.75$, sampling temperature
$\tau=1$, and a coarse noise grid of size $5\times8\times12$.
Each candidate retains $\rho=0.10$ of the spatiotemporal latent positions, and candidate errors are converted into weights using $\beta=0.7$.

\subsection{Backward: Interaction-Weighted Supervision}
\label{sec:iws}

IWS uses the interaction map $\mathbf{G}$ to increase the contribution of interaction-relevant positions during denoising optimization.
For each position $p\in\Omega$, we define
\begin{equation}
    w_p
    =
    1
    +
    (\gamma-1)\operatorname{sg}[G_p],
    \label{eq:interaction_weight}
\end{equation}
where $\gamma\geq1$ controls the maximum regional emphasis.
Because $G_p\in[0,1]$, the resulting weight satisfies
$w_p\in[1,\gamma]$.
Positions outside the estimated interaction region retain their original unit weight, while positions with high interaction-map values receive stronger supervision.

The interaction-weighted denoising objective is
\begin{equation}
    \mathcal{L}_{\mathrm{IWS}}(\theta)
    =
    \mathbb{E}_{\mathbf{z}_0,\boldsymbol{\epsilon},t}
    \left[
    \frac{
    \sum_{p\in\Omega}w_p\ell_p
    }{
    \sum_{p\in\Omega}w_p
    }
    \right].
    \label{eq:iws_objective}
\end{equation}
The normalization by the total weight prevents the loss scale from growing with the size or magnitude of the emphasized region.
Unlike a masked regional loss, \eqnref{eq:iws_objective} retains supervision over the complete spatiotemporal field and only changes its spatial allocation.
We use $\gamma=5$ in the main experiments.

\paragraph{Gradient-decoupled optimization.}
The interaction map is derived from cross-attention, creating a dependency between region estimation and the objective guided by that region.
Although $\mathbf{G}$ is detached, allowing
$\mathcal{L}_{\mathrm{IWS}}$ to update the attention-producing parameters could still cause the cross-attention representation itself to adapt to the regional objective in subsequent training steps.
We therefore separate the DiT parameters into the cross-attention parameters
$\theta_{\mathrm{ca}}$ and all remaining parameters
$\theta_{\mathrm{rest}}$.
Their gradients are routed according to
\begin{equation}
    \nabla_{\theta_{\mathrm{ca}}}
    \mathcal{L}_{\method}
    =
    \nabla_{\theta_{\mathrm{ca}}}
    \mathcal{L}_{\mathrm{global}},
    \label{eq:cross_attention_routing}
\end{equation}
and
\begin{equation}
    \nabla_{\theta_{\mathrm{rest}}}
    \mathcal{L}_{\method}
    =
    \nabla_{\theta_{\mathrm{rest}}}
    \mathcal{L}_{\mathrm{IWS}}.
    \label{eq:remaining_parameter_routing}
\end{equation}
In practice, this routing is implemented through two backward passes with parameter-group gradient hooks.
The global backward pass retains gradients only for
$\theta_{\mathrm{ca}}$, while the IWS backward pass retains gradients only for
$\theta_{\mathrm{rest}}$.
Consequently, the cross-attention used to estimate interaction regions remains governed by the original uniformly supervised objective, whereas the remaining DiT parameters learn from the spatially reallocated supervision.
This separation prevents the regional objective from directly optimizing its own localization signal and decouples interaction-region estimation from interaction-focused optimization.

\paragraph{Training and inference cost.}
\method{} reuses the cross-attention probabilities and token-wise prediction errors already produced during standard world-model training.
Its additional computation consists primarily of sampling $K$ candidate masks, evaluating their masked mean errors, and constructing the interaction-weighted objective.
No component of ADS or IWS is used at inference time, so the trained world model preserves the original architecture and inference procedure.

\section{Experiments}
\label{sec:experiments}

\begin{table*}[t]
\centering
\begin{tabular}{l c cccccc}
\toprule
\multicolumn{1}{c}{\raisebox{0.85ex}{Model}} & \multicolumn{1}{c}{\raisebox{0.85ex}{EWMScore}} & \shortstack{Visual\\Quality} & \shortstack{Motion\\Quality} & \shortstack{Content\\Consist.} & \shortstack{Physics\\Adher.} & \shortstack{3D\\Accuracy} & \shortstack{Control-\\lability} \\
\midrule
\rowcolor{gray!15}\multicolumn{8}{l}{\emph{General world models}} \\
CogVideoX & 57.90 & 55.81 & 42.49 & 68.12 & 47.33 & 84.62 & 54.40 \\
Veo 3.1 & 58.87 & 56.44 & 46.12 & 66.12 & 45.52 & 78.48 & \underline{62.62} \\
Wan 2.6 & 61.86 & \textbf{61.62} & \textbf{68.31} & 60.36 & 42.31 & 75.88 & 60.85 \\
\midrule
\rowcolor{gray!15}\multicolumn{8}{l}{\emph{Embodied world models}} \\
GigaWorld-0 & 53.39 & 44.82 & 58.79 & 58.74 & 34.60 & 69.56 & 52.94 \\
Genie Envisioner & 43.65 & 29.78 & 49.17 & 62.63 & 13.66 & 69.73 & 35.60 \\
Vidar & 51.60 & 46.07 & 40.55 & 60.93 & 36.38 & 77.32 & 51.86 \\
IRASim & 58.12 & 54.81 & 44.25 & \textbf{69.67} & 46.48 & 85.50 & 53.26 \\
CtrlWorld & 59.70 & 55.33 & 50.28 & 63.99 & \underline{54.89} & \underline{86.30} & 54.65 \\
\midrule
\rowcolor{gray!15}\multicolumn{8}{l}{\emph{Representation-guided models}} \\
TesserAct & 53.23 & 41.64 & 50.59 & 66.60 & 35.98 & 75.39 & 50.82 \\
RoboMaster & 51.84 & 34.32 & 48.49 & \underline{69.25} & 32.61 & 79.62 & 49.62 \\
WoW & 54.88 & 52.98 & 50.02 & 64.52 & 38.11 & 74.77 & 49.89 \\
\midrule
\rowcolor{gray!15}\multicolumn{8}{l}{\emph{Wan 2.2-based models}} \\
Wan 2.2 & 50.79 & 51.41 & 42.12 & 54.02 & 34.05 & 77.14 & 49.22 \\
Wan 2.2-AC & 58.65 & 56.57 & 48.23 & 60.55 & 53.41 & 86.16 & 54.40 \\
Wan 2.2-AC + \method{} & \underline{62.46} & \underline{60.60} & 53.28 & 60.29 & \textbf{55.87} & \textbf{92.56} & 60.00 \\
\midrule
\rowcolor{gray!15}\multicolumn{8}{l}{\emph{Cosmos-based models}} \\
Cosmos-Predict 2.5 (text) & 50.81 & 47.65 & 60.32 & 57.94 & 23.44 & 75.08 & 39.38 \\
Cosmos-Predict 2.5 (action) & 55.91 & 57.87 & 38.89 & 68.73 & 42.23 & 82.53 & 49.51 \\
Cosmos-Predict 2.5 (action) + \method{} & \textbf{62.53} & 56.32 & \underline{68.22} & 57.90 & 41.82 & 77.99 & \textbf{71.19} \\
\bottomrule
\end{tabular}
\caption{Generation quality on WorldArena. Models are grouped into general, embodied, and representation-guided models, with Wan~2.2-based and Cosmos-based variants listed separately. We report the overall EWMScore and six aggregate dimensions, each on a $[0,100]$ scale where higher is better ($\uparrow$). Boldface and underlining denote the best and second-best results in each column.}
\label{tab:main}
\end{table*}

\begin{table}[!t]
\centering
\small
\setlength{\tabcolsep}{1mm}
\begin{tabular}{l cc cc}
\toprule
& \multicolumn{2}{c}{Visual} & \multicolumn{2}{c}{Hand interaction} \\
\cmidrule(lr){2-3}\cmidrule(lr){4-5}
Model & FVD $\downarrow$ & FID $\downarrow$ & \shortstack{CLIP-\\Hand $\uparrow$} & \shortstack{Hand\\IoU $\uparrow$} \\
\midrule
\rowcolor{gray!15}\multicolumn{5}{l}{\emph{General world models}} \\
HunyuanVideo-1.5 & 541.83 & 56.73 & 0.902 & 0.328 \\
Cosmos-Predict 2.5 & 615.42 & 50.12 & 0.914 & 0.386 \\
\midrule
\rowcolor{gray!15}\multicolumn{5}{l}{\emph{Pose control}} \\
MimicMotion & 612.75 & 48.55 & 0.882 & 0.492 \\
MagicPose & 1456.20 & 212.94 & 0.864 & 0.298 \\
VACE & \underline{358.42} & 50.65 & 0.895 & 0.493 \\
LOME & 1748.29 & 66.03 & 0.745 & 0.087 \\
\midrule
\rowcolor{gray!15}\multicolumn{5}{l}{\emph{Wan 2.2-based models}} \\
Wan 2.2 & 1463.05 & 199.35 & 0.876 & 0.557 \\
Wan 2.2-AC & 366.12 & \underline{44.71} & \underline{0.921} & \underline{0.693} \\
Wan 2.2-AC + \method{} & \textbf{110.94} & \textbf{5.79} & \textbf{0.952} & \textbf{0.772} \\
\bottomrule
\end{tabular}
\caption{Generation quality on EgoDex. Boldface and underlining denote the best and second-best results in each column.}
\label{tab:egodex}
\end{table}

\subsection{Setup}
\label{sec:setup}

\paragraph{Implementation details.}
We evaluate \method{} in two manipulation settings to demonstrate the generality of our method across interaction types, applying it to the Wan2.2 and Cosmos-Predict 2.5 backbone in both. For robot-arm manipulation, the model is trained on $\sim$350K 17-frame videos collected from \textbf{RoboTwin}~\citep{chen2025robotwin}, conditioned on 14-DoF dual-arm action trajectories, injected through an additional action encoder. For human-hand manipulation, we train on $\sim$256K clips drawn from \textbf{EgoDex}~\citep{hoque2025egodex} across 118 tasks, using 81-frame videos conditioned on hand-pose videos temporally aligned with the RGB frames, injected through the shared VAE encoder without any additional encoder. Both settings are trained at 720p and optimization is identical across the two settings: we use bf16 mixed precision with FSDP over 8 GPUs, a per-device batch size of 1 with 4-step gradient accumulation (effective global batch size $1\times8\times4=32$), a constant learning rate of $2\times10^{-5}$ after 100 warm-up steps, and we train for one epoch. \method{} hyperparameters are fixed across both settings: $K=8$, $\sigma=0.75$, coarse grid $5\times8\times12$, $\rho=0.10$, $\kappa=0.65$, $\beta=0.7$, $\tau=1$, $\gamma=5$.

\paragraph{Benchmarks and metrics.}
For robot-arm manipulation, we evaluate on \textbf{WorldArena} \citep{shang2026worldarena}, which scores dual-arm manipulation along six dimensions: Visual Quality, Motion Quality, Content Consistency, Physics Adherence, 3D Accuracy, and Controllability, spanning 16 normalized metrics, and condenses overall generation quality into a single \textbf{EWMScore} (the mean of the 16 metrics). We report EWMScore together with the six aggregate dimensions in \tabref{tab:main}, and provide all 16 metrics in the technical appendix.
For human-hand manipulation, we evaluate on the \textbf{EgoDex} \citep{hoque2025egodex} test set along two axes (\tabref{tab:egodex}): \emph{visual} metrics: FVD~\citep{unterthiner2018fvd} and FID~\citep{heusel2017gans}, covering temporal coherence and per-frame appearance quality; and \emph{hand-interaction} metrics: CLIP-Hand for the local appearance and semantics of the hand and nearby manipulated object, and Hand IoU for the coarse 2D position and scale of the generated hand~\citep{sun2026handworld}.

\paragraph{Baselines.}
For robot-arm manipulation, all models follow the WorldArena evaluation settings. Existing baselines comprise general world models: CogVideoX~\citep{yang2024cogvideox}, Wan~2.6~\citep{wan2025wan}, and Veo~3.1; embodied world models: GigaWorld-0, Genie Envisioner, Vidar, IRASim~\citep{zhu2025irasim}, and CtrlWorld; and representation-guided models: TesserAct~\citep{zhen2025tesseract}, RoboMaster, and WoW. We additionally report two backbone-specific groups to evaluate \method{}: Wan~2.2-based models include Wan~2.2~\citep{wan2025wan}, MSE-trained Wan~2.2-AC, and Wan~2.2-AC with \method{}; Cosmos-based models include Cosmos-Predict~2.5 (text)~\citep{agarwal2025cosmos}, Cosmos-Predict~2.5 (action), and its \method{} variant. For human-hand manipulation, baselines comprise general video world models: HunyuanVideo-1.5~\citep{wu2025hunyuanvideo15technicalreport} and Cosmos-Predict~2.5~\citep{agarwal2025cosmos} and pose-controlled models: MimicMotion~\citep{zhang2025mimicmotionhighqualityhumanmotion}, MagicPose~\citep{chang2024magicposerealistichumanposes}, VACE~\citep{jiang2025vace}, and LOME~\citep{gao2026lomelearninghumanobjectmanipulation}.

\begin{figure*}[!t]
\centering
\includegraphics[width=\linewidth]{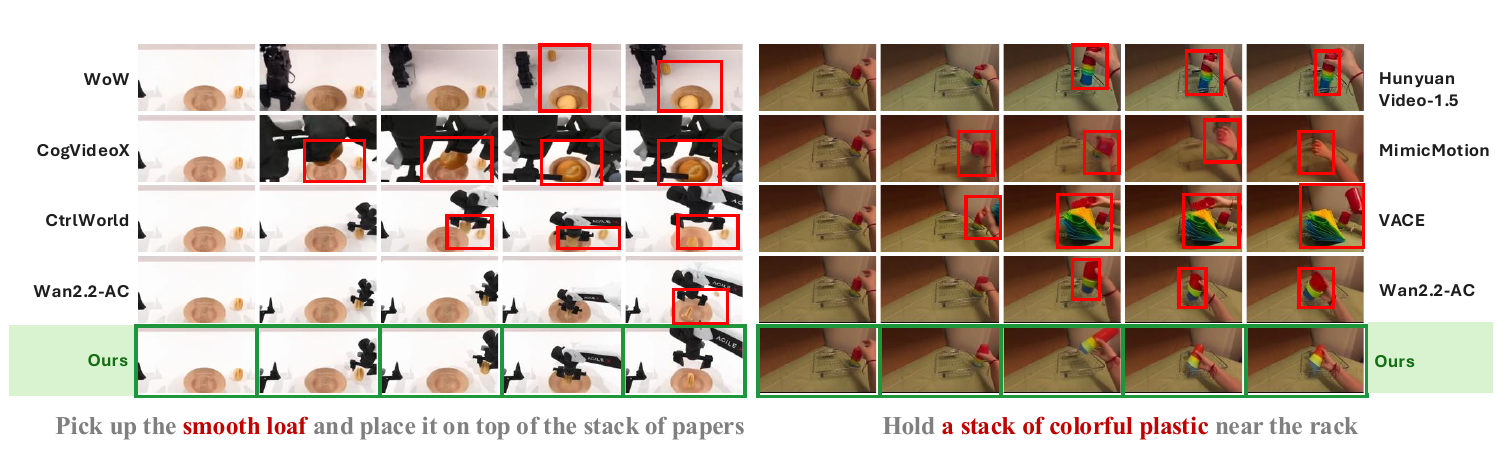}
\caption{Qualitative comparison on both interaction settings. Left: robot-arm manipulation on WorldArena; right: first-person human-hand manipulation on EgoDex. For each setting we show generated frames for a representative instruction against various baselines. In both settings, \method{} follows the instruction more faithfully and renders the interaction. Red boxes highlight artifacts and inconsistencies in the interaction regions of the baseline generations.}
\label{fig:qualitative}

\includegraphics[width=\linewidth]{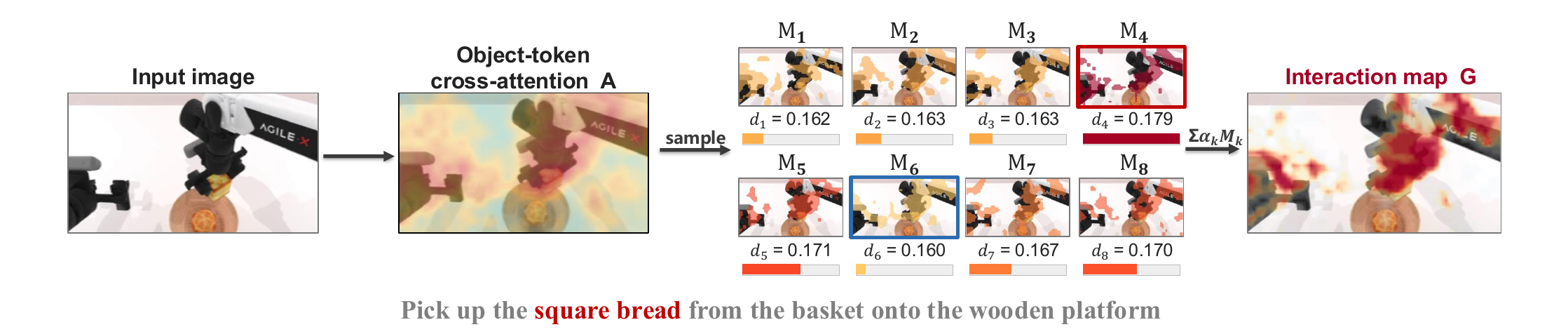}
\caption{Visualization of ADS. On one robot-arm training forward pass, the light-to-dark color bar encodes increasing candidate errors and thus the relative contribution of each candidate to the aggregate.}
\label{fig:ads_calibration}
\end{figure*}

\subsection{Quantitative Analysis}
\label{sec:quantitative_analysis}

\paragraph{Robot-arm manipulation.}
\tabref{tab:main} shows that \method{} improves both backbone families. On Cosmos-Predict~2.5 (action), it raises EWMScore from 55.91 to 62.53 (+6.62 points, 11.8\%), achieving the best overall result, the best Controllability (71.19), and the second-best Motion Quality (68.22). Interaction Quality increases from 0.5500 to 0.6360 and Action Following from 0.0133 to 0.6260, although the remaining aggregate dimensions decline. On Wan~2.2-AC, \method{} raises EWMScore from 58.65 to 62.46 (+3.81 points, 6.5\%) and improves Visual Quality, Motion Quality, Physics Adherence, 3D Accuracy, and Controllability, attaining the best Physics Adherence (55.87) and 3D Accuracy (92.56) and the second-best Visual Quality (60.60). The best \method{} result exceeds Wan~2.6, CtrlWorld, and WoW by 0.67, 2.83, and 7.65 points, respectively.

\paragraph{Human-hand manipulation.}
For human-hand manipulation, \tabref{tab:egodex} reports that \method{} attains the best results on  visual metrics across both general video world models and pose-controlled models, cutting FVD from 366.12 to 110.94 and FID from 44.71 to 5.79 over the action-conditioned Wan~2.2-AC on the same backbone.
It also leads on both hand-interaction metrics, improving CLIP-Hand from 0.921 to 0.952 and Hand IoU from 0.693 to 0.772 over Wan~2.2-AC on the same backbone. These gains indicate stronger local interaction fidelity and hand localization than both the MSE-trained counterpart and the pose-controlled baselines. Together with the robot-arm results, these improvements demonstrate that \method{} delivers consistent gains across different DiT backbones and control signals.

\subsection{Qualitative analysis}
\label{sec:Qualitative analysis}
\paragraph{Generation comparison.}
\figref{fig:qualitative} compares generations in robot-arm and human-hand manipulation. In both cases, the baselines follow the instruction loosely and tend to blur or distort the contact region, whether the gripper--object contact for the robot arm or the hand--object contact for the human hand, and some drift away from the target. In contrast, \method{} produces the specified interaction with a sharper contact region and more coherent object dynamics, while keeping the surrounding scene stable. This clearly demonstrates the effectiveness of \method{} for interaction-region generation.

\paragraph{ADS calibration.}
\figref{fig:ads_calibration} illustrates how ADS calibrates the attention prior in a representative training example. The raw object-conditioned cross-attention map $\mathbf{A}$ is diffuse, spreading across the robot arm, workspace, and background. ADS samples $K{=}8$ candidate regions $\mathbf{M}_1,\dots,\mathbf{M}_8$ from this prior and evaluates them using detached local prediction errors (\eqsref{eq:candidate_score}). Candidates covering the contact region receive higher weights than those dominated by the static background. Their weighted aggregation (\eqsref{eq:interaction_map}) produces an interaction map $\mathbf{G}$ that is more concentrated around the interacting arm, gripper, and manipulated object. This example illustrates how ADS refines a coarse attention prior into a more targeted supervision map for IWS.

\subsection{Ablation Studies}
\label{sec:ablation}

\paragraph{Component ablation: IWS and ADS.}
\tabref{tab:ads_sampling} shows the effect of separating the components on WorldArena and reveals their complementarity. Starting from the AC backbone (58.65 EWMScore), IWS provides the more direct gain (+2.89 to 61.54), since it primarily addresses the supervision-allocation mismatch. On top of this, ADS further calibrates the cross-attention prior and delivers an additional improvement (+0.92 to 62.46) over weighting the raw, coarse prior directly. Integrated within a single training step, the two components jointly strengthen interaction-region generation, improving EWMScore by 3.81 points overall.

\begin{table}[!t]
\centering
\small
\setlength{\tabcolsep}{1mm}
\begin{tabular}{l cccc}
\toprule
\multicolumn{1}{c}{\raisebox{0.85ex}{Method}} & \multicolumn{1}{c}{\raisebox{0.85ex}{EWMScore}} & \shortstack{Visual\\Quality} & \shortstack{Motion\\Quality} & \shortstack{Physics\\Adher.} \\
\midrule
Wan 2.2-AC & 58.65 & 56.57 & 48.23 & 53.41 \\
\quad + IWS & 61.54 & 60.44 & 49.16 & 55.16 \\
\quad + IWS + ADS (\method{}) & \textbf{62.46} & \textbf{60.60} & \textbf{53.28} & \textbf{55.87} \\
\bottomrule
\end{tabular}
\caption{Per-component ablation on WorldArena.}
\label{tab:ads_sampling}
\end{table}

\section{Conclusion}
\label{sec:conclusion}

In this work, we introduced \method{}, a scalable framework that addresses the supervision-allocation mismatch by converting object-conditioned cross-attention into targeted denoising supervision. ADS calibrates attention proposals with detached local prediction errors, while IWS reweights training with the resulting interaction map, requiring neither external spatial signals nor inference-time changes. Experiments on robot-arm and human-hand manipulation show consistent gains over uniform MSE training and strong baselines, demonstrating the effectiveness and scalability of \method{} for interaction-aware world model training.

\bibliography{impact_aaai27}

@article{bruce2024genie,
  title = {Genie: Generative Interactive Environments},
  author = {Bruce, Jake and Dennis, Michael and Edwards, Ashley and Parker-Holder, Jack and Shi, Yuge and Hughes, Edward and Lai, Matthew and Mavalankar, Aditi and Steigerwald, Richie and Apps, Chris and Aytar, Yusuf and Bechtle, Sarah and Behbahani, Feryal and Chan, Stephanie C. Y. and Heess, Nicolas and Gonzalez, Lucy and Osindero, Simon and Ozair, Sherjil and Reed, Scott and Zhang, Jingwei and Zolna, Konrad and Clune, Jeff and Freitas, Nando de and Rocktaschel, Tim and Hafner, Danijar},
  journal = {arXiv preprint arXiv:2402.15391},
  year = {2024}
}

@inproceedings{heusel2017gans,
  title = {GANs Trained by a Two Time-Scale Update Rule Converge to a Local Nash Equilibrium},
  author = {Heusel, Martin and Ramsauer, Hubert and Unterthiner, Thomas and Nessler, Bernhard and Hochreiter, Sepp},
  booktitle = {Advances in Neural Information Processing Systems},
  year = {2017}
}

@article{unterthiner2018fvd,
  title = {Towards Accurate Generative Models of Video: A New Metric and Challenges},
  author = {Unterthiner, Thomas and van Steenkiste, Sjoerd and Kurach, Karol and Marinier, Raphael and Michalski, Marcin and Gelly, Sylvain},
  journal = {arXiv preprint arXiv:1812.01717},
  year = {2018}
}

@inproceedings{lipman2023flow,
  title     = {Flow Matching for Generative Modeling},
  author    = {Lipman, Yaron and Chen, Ricky T. Q. and Ben-Hamu, Heli and Nickel, Maximilian and Le, Matt},
  booktitle = {International Conference on Learning Representations},
  year      = {2023}
}

@article{yang2024cogvideox,
  title   = {{CogVideoX}: Text-to-Video Diffusion Models with An Expert Transformer},
  author  = {Yang, Zhuoyi and Teng, Jiayan and Zheng, Wendi and Ding, Ming and Huang, Shiyu and Xu, Jiazheng and Yang, Yuanming and Hong, Wenyi and Zhang, Xiaohan and Feng, Guanyu and Yin, Da and Gu, Xiaotao and Zhang, Yuxuan and Wang, Weihan and Cheng, Yean and Liu, Ting and Xu, Bin and Dong, Yuxiao and Tang, Jie},
  journal = {arXiv preprint arXiv:2408.06072},
  year    = {2024}
}

@article{kong2024hunyuanvideo,
  title   = {{HunyuanVideo}: A Systematic Framework For Large Video Generative Models},
  author  = {Kong, Weijie and Tian, Qi and Zhang, Zijian and Min, Rox and Dai, Zuozhuo and Zhou, Jin and Xiong, Jiangfeng and Li, Xin and Wu, Bo and Zhang, Jianwei and others},
  journal = {arXiv preprint arXiv:2412.03603},
  year    = {2024}
}

@article{wan2025wan,
  title   = {Wan: Open and Advanced Large-Scale Video Generative Models},
  author  = {{Wan Team} and Wang, Ang and Ai, Baole and Wen, Bin and Mao, Chaojie and Xie, Chen-Wei and Chen, Di and Yu, Feiwu and Zhao, Haiming and Yang, Jianxiao and others},
  journal = {arXiv preprint arXiv:2503.20314},
  year    = {2025}
}

@article{agarwal2025cosmos,
  title   = {Cosmos World Foundation Model Platform for Physical AI},
  author  = {{NVIDIA} and Agarwal, Niket and Ali, Arslan and Bala, Maciej and Balaji, Yogesh and Barker, Erik and Cai, Tiffany and Chattopadhyay, Prithvijit and Chen, Yongxin and Cui, Yin and others},
  journal = {arXiv preprint arXiv:2501.03575},
  year    = {2025}
}

@article{chen2025robotwin,
  title   = {{RoboTwin} 2.0: A Scalable Data Generator and Benchmark with Strong Domain Randomization for Robust Bimanual Robotic Manipulation},
  author  = {Chen, Tianxing and Chen, Zanxin and Chen, Baijun and Cai, Zijian and Liu, Yibin and Li, Zixuan and Liang, Qiwei and Lin, Xianliang and Ge, Yiheng and Gu, Zhenyu and Deng, Weiliang and Guo, Yubin and Nian, Tian and Xie, Xuanbing and Chen, Qiangyu and Su, Kailun and Xu, Tianling and Liu, Guodong and Hu, Mengkang and Gao, Huan-ang and Wang, Kaixuan and Liang, Zhixuan and Qin, Yusen and Yang, Xiaokang and Luo, Ping and Mu, Yao},
  journal = {arXiv preprint arXiv:2506.18088},
  year    = {2025}
}

@misc{hoque2025egodex,
  title         = {{EgoDex}: Learning Dexterous Manipulation from Large-Scale Egocentric Video},
  author        = {Hoque, Ryan and Huang, Peide and Yoon, David J. and Sivapurapu, Mouli and Zhang, Jian},
  year          = {2025},
  eprint        = {2505.11709},
  archivePrefix = {arXiv},
  primaryClass  = {cs.CV},
  url           = {https://arxiv.org/abs/2505.11709}
}

@inproceedings{sun2026handworld,
  title     = {HandWorld: Hand-Centric Unified Video Action Generation},
  author    = {Sun, Zhihao and Du, Zhiying and Yang, Xitong and Wu, Zuxuan},
  booktitle = {Proceedings of the IEEE/CVF Conference on Computer Vision and Pattern Recognition},
  pages     = {15976--15985},
  year      = {2026}
}

@article{jiang2025vace,
  title   = {{VACE}: All-in-One Video Creation and Editing},
  author  = {Jiang, Zeyinzi and Han, Zhen and Mao, Chaojie and Zhang, Jingfeng and Pan, Yulin and Liu, Yu},
  journal = {arXiv preprint arXiv:2503.07598},
  year    = {2025}
}

@article{wu2026ltd,
  title   = {Latent Temporal Discrepancy as Motion Prior: A Loss-Weighting Strategy for Dynamic Fidelity in T2V},
  author  = {Wu, Meiqi and Song, Bingze and Lin, Ruimin and Zhu, Chen and Feng, Xiaokun and Wu, Jiahong and Chu, Xiangxiang and Huang, Kaiqi},
  journal = {arXiv preprint arXiv:2601.20504},
  year    = {2026}
}

@article{tian2026starry,
  title   = {{STARRY}: Spatial-Temporal Action-Centric World Modeling for Robotic Manipulation},
  author  = {Tian, Yuxuan and Jin, Yurun and Yu, Bin and Shi, Yukun and Wu, Hao and Liu, Chi Harold and Chen, Kai and Huang, Cong},
  journal = {arXiv preprint arXiv:2604.26848},
  year    = {2026}
}

@article{yang2026eawm,
  title   = {{EA-WM}: Event-Aware Generative World Model with Structured Kinematic-to-Visual Action Fields},
  author  = {Yang, Zhaoyang and Jin, Yurun and Qi, Lizhe and Huang, Cong and Chen, Kai},
  journal = {arXiv preprint arXiv:2605.06192},
  year    = {2026}
}

@article{fang2025motionimage,
  title   = {Robotic {VLA} Benefits from Joint Learning with Motion Image Diffusion},
  author  = {Fang, Yu and Ranasinghe, Kanchana and Xue, Le and Zhou, Honglu and Tan, Juntao and Xu, Ran and Heinecke, Shelby and Xiong, Caiming and Savarese, Silvio and Szafir, Daniel and Ding, Mingyu and Ryoo, Michael S. and Niebles, Juan Carlos},
  journal = {arXiv preprint arXiv:2512.18007},
  year    = {2025}
}

@article{liu2025geometryaware,
  title   = {Geometry-Aware 4D Video Generation for Robot Manipulation},
  author  = {Liu, Zeyi and Li, Shuang and Cousineau, Eric and Feng, Siyuan and Burchfiel, Benjamin and Song, Shuran},
  journal = {arXiv preprint arXiv:2507.01099},
  year    = {2025}
}

@article{yan2025scar,
  title   = {Open-World Hand-Object Interaction Video Generation Based on Structure and Contact-Aware Representation},
  author  = {Yan, Haodong and Yu, Hang and Zhong, Zhide and Yuan, Weilin and Gong, Xin and Luo, Zehang and Heyu, Chengxi and Li, Junfeng and Song, Wenxuan and Zhou, Shunbo and Li, Haoang},
  journal = {arXiv preprint arXiv:2512.01677},
  year    = {2025}
}

@article{luo2026cointeract,
  title   = {{CoInteract}: Physically-Consistent Human-Object Interaction Video Synthesis via Spatially-Structured Co-Generation},
  author  = {Luo, Xiangyang and Xin, Xiaozhe and Feng, Tao and Guo, Xu and Jin, Meiguang and Ma, Junfeng},
  journal = {arXiv preprint arXiv:2604.19636},
  year    = {2026}
}

@article{qwen2025qwen25,
  title   = {{Qwen2.5} Technical Report},
  author  = {{Qwen} and Yang, An and Yang, Baosong and Zhang, Beichen and Hui, Binyuan and Zheng, Bo and Yu, Bowen and Li, Chengyuan and Liu, Dayiheng and Huang, Fei and Wei, Haoran and Lin, Huan and Yang, Jian and Tu, Jianhong and Zhang, Jianwei and Yang, Jianxin and Yang, Jiaxi and Zhou, Jingren and Lin, Junyang and Dang, Kai and Lu, Keming and Bao, Keqin and Yang, Kexin and Yu, Le and Li, Mei and Xue, Mingfeng and Zhang, Pei and Zhu, Qin and Men, Rui and Lin, Runji and Li, Tianhao and Tang, Tianyi and Xia, Tingyu and Ren, Xingzhang and Ren, Xuancheng and Fan, Yang and Su, Yang and Zhang, Yichang and Wan, Yu and Liu, Yuqiong and Cui, Zeyu and Zhang, Zhenru and Qiu, Zihan},
  journal = {arXiv preprint arXiv:2412.15115},
  year    = {2025},
  url     = {https://arxiv.org/abs/2412.15115}
}

@misc{alayaworld2026,
  title         = {{AlayaWorld}: Long-Horizon and Playable Video World Generation},
  author        = {{AlayaWorld Team} and Zhang, Kaipeng and Li, Chuanhao and Zhan, Yifan and Ge, Yongtao and Yin, Yuanyang and Tan, Jiaming and He, Kang and Fan, Liaoyuan and Liu, Ruicong and Xu, Xiaojie and Chu, Xuangeng and Li, Zhen and Lin, Zhengyuan and Wang, Zhixiang and Meng, Zian and Gao, Zihui},
  year          = {2026},
  eprint        = {2607.06291},
  archivePrefix = {arXiv},
  url           = {https://arxiv.org/abs/2607.06291}
}

@misc{hu2026multiplayer,
  title         = {Multiplayer Interactive World Models with Representation Autoencoders},
  author        = {Hu, Anthony and Volhejn, V{\'a}clav and Rahary, Adrien Ramanana and Mulder, Chris and Makkar, Aditya and Royer, Am{\'e}lie and Orsini, Manu and Liao, Alyx and Jelley, Adam and Alonso, Eloi and Laurent, Florian and Nor{\'e}n, Fredrik and Swingos, James and H{\"u}nermann, Jan and Rollins, Kent and Hosseini, Lucas and Le Cauchois, Matthieu and Peter, Maxim and de Witte, Pim and Brown, Tim and Micheli, Vincent and B{\"o}hle, Moritz and de Marmiesse, Gabriel and Sharmanska, Viktoriia and Specia, Lucia and Black, Michael and P{\'e}rez, Patrick},
  year          = {2026},
  eprint        = {2607.05352},
  archivePrefix = {arXiv},
  url           = {https://arxiv.org/abs/2607.05352}
}

@misc{zhang2026qwenrobotworld,
  title         = {{Qwen-RobotWorld} Technical Report: Unifying Embodied World Modeling through Language-Conditioned Video Generation},
  author        = {Zhang, Jie and Chen, Xiaoyue and Chen, Anzhe and Lv, Chenxu and Li, Deqing and Zhou, Gengze and Yin, Hang and Yuan, Haoqi and Li, Haoyang and Li, Jiahao and Zhang, Jiazhao and Zhou, Jingren and Gao, Kaiyuan and Yan, Kun and Jiang, Lihan and Tang, Ningyuan and Lin, Pei and Peng, Qihang and Yin, Shengming and Wu, Tianhe and Yan, Tianyi and Xu, Xiao and Shu, Yan and Zhang, Yanran and Wang, Ye and Wang, Yi and Chen, Yilei and Xu, Yixian and Huang, Yiyang and Chen, Yuxiang and Zhang, Zekai and Wang, Zhendong and Lei, Zhixing and Liang, Zhixuan and Liu, Zihao and Zhou, Zikai and Chen, Xiong-Hui and Wu, Chenfei},
  year          = {2026},
  eprint        = {2606.17030},
  archivePrefix = {arXiv},
  url           = {https://arxiv.org/abs/2606.17030}
}

@inproceedings{bi2026motus,
  title     = {{Motus}: A Unified Latent Action World Model},
  author    = {Bi, Hongzhe and Tan, Hengkai and Xie, Shenghao and Wang, Zeyuan and Huang, Shuhe and Liu, Haitian and Zhao, Ruowen and Feng, Yao and Xiang, Chendong and Rong, Yinze and Zhao, Hongyan and Liu, Hanyu and Su, Zhizhong and Ma, Lei and Su, Hang and Zhu, Jun},
  booktitle = {Proceedings of the IEEE/CVF Conference on Computer Vision and Pattern Recognition},
  pages     = {35101--35113},
  year      = {2026},
  url       = {https://openaccess.thecvf.com/content/CVPR2026/html/Bi_Motus_A_Unified_Latent_Action_World_Model_CVPR_2026_paper.html}
}

@misc{zhang2026physisforcing,
  title         = {{PhysisForcing}: Physics Reinforced World Simulator for Robotic Manipulation},
  author        = {Zhang, Peiwen and Deng, Yufan and Sun, Shangkun and Ma, Juncheng and Wang, Duomin and Du, Jonas and Pan, Zilin and Huang, Ye and Liang, Hao and Huang, Songyan and Zhang, Ruihua and Xie, Enze and Liu, Ming-Yu and Zhou, Daquan},
  year          = {2026},
  eprint        = {2606.28128},
  archivePrefix = {arXiv},
  primaryClass  = {cs.CV},
  url           = {https://arxiv.org/abs/2606.28128}
}

@misc{wang2026lifting,
  title         = {Lifting Embodied World Models for Planning and Control},
  author        = {Wang, Alex N. and Darrell, Trevor and Izmailov, Pavel and Bai, Yutong and Bar, Amir},
  year          = {2026},
  eprint        = {2604.26182},
  archivePrefix = {arXiv},
  url           = {https://arxiv.org/abs/2604.26182}
}

@inproceedings{gao2025flip,
  title     = {{FLIP}: Flow-Centric Generative Planning as General-Purpose Manipulation World Model},
  author    = {Gao, Chongkai and Zhang, Haozhuo and Xu, Zhixuan and Cai, Zhehao and Shao, Lin},
  booktitle = {International Conference on Learning Representations},
  year      = {2025},
  url       = {https://proceedings.iclr.cc/paper_files/paper/2025/hash/37bf751554eb05465ed5dc60dced3f38-Abstract-Conference.html}
}

@inproceedings{hu2025vpp,
  title     = {Video Prediction Policy: A Generalist Robot Policy with Predictive Visual Representations},
  author    = {Hu, Yucheng and Guo, Yanjiang and Wang, Pengchao and Chen, Xiaoyu and Wang, Yen-Jen and Zhang, Jianke and Sreenath, Koushil and Lu, Chaochao and Chen, Jianyu},
  booktitle = {Proceedings of the 42nd International Conference on Machine Learning},
  pages     = {24328--24346},
  volume    = {267},
  series    = {Proceedings of Machine Learning Research},
  publisher = {PMLR},
  year      = {2025},
  url       = {https://proceedings.mlr.press/v267/hu25g.html}
}

@inproceedings{zhen2025tesseract,
  title     = {Learning 4D Embodied World Models},
  author    = {Zhen, Haoyu and Sun, Qiao and Zhang, Hongxin and Li, Junyan and Zhou, Siyuan and Du, Yilun and Gan, Chuang},
  booktitle = {Proceedings of the IEEE/CVF International Conference on Computer Vision},
  pages     = {5337--5347},
  year      = {2025},
  url       = {https://openaccess.thecvf.com/content/ICCV2025/html/Zhen_Learning_4D_Embodied_World_Models_ICCV_2025_paper.html}
}

@misc{shang2026worldarena,
  title         = {{WorldArena}: A Unified Benchmark for Evaluating Perception and Functional Utility of Embodied World Models},
  author        = {Shang, Yu and Li, Zhuohang and Ma, Yiding and Su, Weikang and Jin, Xin and Wang, Ziyou and Jin, Lei and Zhang, Xin and Tang, Yinzhou and Su, Haisheng and Gao, Chen and Wu, Wei and Liu, Xihui and Shah, Dhruv and Zhang, Zhaoxiang and Chen, Zhibo and Zhu, Jun and Tian, Yonghong and Chua, Tat-Seng and Zhu, Wenwu and Li, Yong},
  year          = {2026},
  eprint        = {2602.08971},
  archivePrefix = {arXiv},
  primaryClass  = {cs.CV},
  url           = {https://arxiv.org/abs/2602.08971}
}

@inproceedings{po2025longcontext,
  author    = {Po, Ryan and Nitzan, Yotam and Zhang, Richard and Chen, Berlin and Dao, Tri and Shechtman, Eli and Wetzstein, Gordon and Huang, Xun},
  title     = {Long-Context State-Space Video World Models},
  booktitle = {Proceedings of the IEEE/CVF International Conference on Computer Vision (ICCV)},
  pages     = {8733--8744},
  month     = {October},
  year      = {2025},
  url       = {https://openaccess.thecvf.com/content/ICCV2025/html/Po_Long-Context_State-Space_Video_World_Models_ICCV_2025_paper.html}
}

@inproceedings{zhu2025irasim,
  author    = {Zhu, Fangqi and Wu, Hongtao and Guo, Song and Liu, Yuxiao and Cheang, Chilam and Kong, Tao},
  title     = {{IRASim}: A Fine-Grained World Model for Robot Manipulation},
  booktitle = {Proceedings of the IEEE/CVF International Conference on Computer Vision (ICCV)},
  pages     = {9834--9844},
  month     = {October},
  year      = {2025},
  url       = {https://openaccess.thecvf.com/content/ICCV2025/html/Zhu_IRASim_A_Fine-Grained_World_Model_for_Robot_Manipulation_ICCV_2025_paper.html}
}

@inproceedings{kim2026dwm,
  author    = {Kim, Byungjun and Kim, Taeksoo and Lee, Junyoung and Joo, Hanbyul},
  title     = {Dexterous World Models},
  booktitle = {Proceedings of the IEEE/CVF Conference on Computer Vision and Pattern Recognition (CVPR)},
  pages     = {29663--29673},
  month     = {June},
  year      = {2026},
  url       = {https://openaccess.thecvf.com/content/CVPR2026/html/Kim_Dexterous_World_Models_CVPR_2026_paper.html}
}

@misc{fang2026worldscape,
  title         = {{Worldscape-MoE}: A Unified Mixture-of-Experts World Model for Scalable Heterogeneous Action Control},
  author        = {Fang, Jianjie and Xu, Yongyan and Wang, Ziyou and Gao, Chen and Huang, Yuchao and Wang, Zhaolu and Tang, Rongze and Jia, Mingyuan and Zhao, Baining and Zhang, Weichen and Zhang, Xin and Su, Haisheng and Shang, Yu and Wu, Wei and Chen, Xinlei and Li, Yong},
  year          = {2026},
  eprint        = {2607.03964},
  archivePrefix = {arXiv},
  primaryClass  = {cs.RO},
  url           = {https://arxiv.org/abs/2607.03964}
}

@misc{su2026worldscape,
  title         = {{WorldScape Policy 2.0}: Empowering Steerable World Action Modeling with Reasoning-Augmented Memory},
  author        = {Su, Haisheng and Liu, Zongdai and Jin, Xin and Dou, Haoxuan and Hu, Chengming and Li, Baorun and Liu, Zhanwang and Xu, Ruiyan and Fang, Jianjie and Zhang, Xin and Yang, Zhenjie and Yang, Xue and Gao, Chen and Yan, Junchi and Li, Yong and Wu, Wei},
  year          = {2026},
  eprint        = {2607.18840},
  archivePrefix = {arXiv},
  primaryClass  = {cs.RO},
  url           = {https://arxiv.org/abs/2607.18840}
}

@misc{zhao2025airscape,
  title         = {{AirScape}: An Aerial Generative World Model with Motion Controllability},
  author        = {Zhao, Baining and Tang, Rongze and Jia, Mingyuan and Wang, Ziyou and Man, Fanghang and Zhang, Xin and Shang, Yu and Zhang, Weichen and Wu, Wei and Gao, Chen and Chen, Xinlei and Li, Yong},
  year          = {2025},
  eprint        = {2507.08885},
  archivePrefix = {arXiv},
  primaryClass  = {cs.RO},
  url           = {https://arxiv.org/abs/2507.08885}
}

@misc{zhao2026worldvln,
  title         = {{WorldVLN}: Autoregressive World Action Model for Aerial Vision-Language Navigation},
  author        = {Zhao, Baining and Xu, Jiacheng and Feng, Weicheng and Zhang, Xin and Wang, Zhaolu and Wang, Haoyang and Ji, Shilong and Wang, Ziyou and Fang, Jianjie and Zheng, Zhiheng and Zhang, Weichen and Shang, Yu and Wu, Wei and Gao, Chen and Chen, Xinlei and Li, Yong},
  year          = {2026},
  eprint        = {2605.15964},
  archivePrefix = {arXiv},
  primaryClass  = {cs.RO},
  url           = {https://arxiv.org/abs/2605.15964}
}

@misc{fang2026iworld,
  title         = {{iWorld-Bench}: A Benchmark for Interactive World Models with a Unified Action Generation Framework},
  author        = {Fang, Jianjie and Lei, Yingshan and Wan, Qin and Wang, Ziyou and Huang, Yuchao and Xu, Yongyan and Zhao, Baining and Zhang, Weichen and Gao, Chen and Chen, Xinlei and Li, Yong},
  year          = {2026},
  eprint        = {2605.03941},
  archivePrefix = {arXiv},
  url           = {https://arxiv.org/abs/2605.03941}
}

@misc{wu2025hunyuanvideo15technicalreport,
  title         = {{HunyuanVideo 1.5} Technical Report},
  author        = {Wu, Bing and Zou, Chang and Li, Changlin and Huang, Duojun and Yang, Fang and Tan, Hao and Peng, Jack and Wu, Jianbing and Xiong, Jiangfeng and Jiang, Jie and {{Linus}} and {{Patrol}} and Zhang, Peizhen and Chen, Peng and Zhao, Penghao and Tian, Qi and Liu, Songtao and Kong, Weijie and Wang, Weiyan and He, Xiao and Li, Xin and Deng, Xinchi and Zhe, Xuefei and Li, Yang and Long, Yanxin and Peng, Yuanbo and Wu, Yue and Liu, Yuhong and Wang, Zhenyu and Dai, Zuozhuo and Peng, Bo and Li, Coopers and Gong, Gu and Xiao, Guojian and Tian, Jiahe and Lin, Jiaxin and Liu, Jie and Zhang, Jihong and Lian, Jiesong and Pan, Kaihang and Wang, Lei and Niu, Lin and Chen, Mingtao and Chen, Mingyang and Zheng, Mingzhe and Yang, Miles and Hu, Qiangqiang and Yang, Qi and Xiao, Qiuyong and Wu, Runzhou and Xu, Ryan and Yuan, Rui and Sang, Shanshan and Huang, Shisheng and Gong, Siruis and Huang, Shuo and Guo, Weiting and Yuan, Xiang and Chen, Xiaojia and Hu, Xiawei and Sun, Wenzhi and Wu, Xiele and Ren, Xianshun and Yuan, Xiaoyan and Mi, Xiaoyue and Zhang, Yepeng and Sun, Yifu and Lu, Yiting and Li, Yitong and Huang, You and Tang, Yu and Li, Yixuan and Deng, Yuhang and Zhou, Yuan and Hu, Zhichao and Liu, Zhiguang and Yang, Zhihe and Yang, Zilin and Lu, Zhenzhi and Zhou, Zixiang and Zhong, Zhao},
  year          = {2025},
  eprint        = {2511.18870},
  archivePrefix = {arXiv},
  primaryClass  = {cs.CV},
  url           = {https://arxiv.org/abs/2511.18870}
}

@inproceedings{zhang2025mimicmotionhighqualityhumanmotion,
  title     = {{MimicMotion}: High-Quality Human Motion Video Generation with Confidence-aware Pose Guidance},
  author    = {Zhang, Yuang and Gu, Jiaxi and Wang, Li-Wen and Wang, Han and Cheng, Junqi and Zhu, Yuefeng and Zou, Fangyuan},
  booktitle = {Proceedings of the 42nd International Conference on Machine Learning},
  pages     = {74896--74910},
  volume    = {267},
  series    = {Proceedings of Machine Learning Research},
  publisher = {PMLR},
  year      = {2025},
  url       = {https://proceedings.mlr.press/v267/zhang25v.html}
}

@inproceedings{chang2024magicposerealistichumanposes,
  title     = {{MagicPose}: Realistic Human Poses and Facial Expressions Retargeting with Identity-aware Diffusion},
  author    = {Chang, Di and Shi, Yichun and Gao, Quankai and Xu, Hongyi and Fu, Jessica and Song, Guoxian and Yan, Qing and Zhu, Yizhe and Yang, Xiao and Soleymani, Mohammad},
  booktitle = {Proceedings of the 41st International Conference on Machine Learning},
  pages     = {6263--6285},
  volume    = {235},
  series    = {Proceedings of Machine Learning Research},
  publisher = {PMLR},
  year      = {2024},
  url       = {https://proceedings.mlr.press/v235/chang24d.html}
}

@misc{gao2026lomelearninghumanobjectmanipulation,
  title         = {{LOME}: Learning Human-Object Manipulation with Action-Conditioned Egocentric World Model},
  author        = {Gao, Quankai and Yang, Jiawei and Xu, Qiangeng and Chen, Le and Wang, Yue},
  year          = {2026},
  eprint        = {2603.27449},
  archivePrefix = {arXiv},
  primaryClass  = {cs.CV},
  url           = {https://arxiv.org/abs/2603.27449}
}

@misc{xiang2024pandora,
  title         = {{Pandora}: Towards General World Model with Natural Language Actions and Video States},
  author        = {Xiang, Jiannan and Liu, Guangyi and Gu, Yi and Gao, Qiyue and Ning, Yuting and Zha, Yuheng and Feng, Zeyu and Tao, Tianhua and Hao, Shibo and Shi, Yemin and Liu, Zhengzhong and Xing, Eric P. and Hu, Zhiting},
  year          = {2024},
  eprint        = {2406.09455},
  archivePrefix = {arXiv},
  primaryClass  = {cs.AI},
  url           = {https://arxiv.org/abs/2406.09455}
}

@misc{he2025matrixgame2,
  title         = {{Matrix-Game 2.0}: An Open-Source, Real-Time, and Streaming Interactive World Model},
  author        = {He, Xianglong and Peng, Chunli and Liu, Zexiang and Wang, Boyang and Zhang, Yifan and Cui, Qi and Kang, Fei and Jiang, Biao and An, Mengyin and Ren, Yangyang and Xu, Baixin and Guo, Hao-Xiang and Gong, Kaixiong and Wu, Cyrus and Li, Wei and Song, Xuchen and Liu, Yang and Li, Eric and Zhou, Yahui},
  year          = {2025},
  eprint        = {2508.13009},
  archivePrefix = {arXiv},
  primaryClass  = {cs.CV},
  url           = {https://arxiv.org/abs/2508.13009}
}

@misc{wang2026hand2world,
  title         = {{Hand2World}: Autoregressive Egocentric Interaction Generation via Free-Space Hand Gestures},
  author        = {Wang, Yuxi and Ouyang, Wenqi and Wei, Tianyi and Dong, Yi and Shen, Zhiqi and Pan, Xingang},
  year          = {2026},
  eprint        = {2602.09600},
  archivePrefix = {arXiv},
  primaryClass  = {cs.CV},
  url           = {https://arxiv.org/abs/2602.09600}
}
\clearpage
\makeatletter
\setlength{\@fptop}{0pt}
\setlength{\@fpsep}{12pt plus 1fil}
\setlength{\@fpbot}{0pt plus 1fil}
\setlength{\@dblfptop}{0pt}
\setlength{\@dblfpsep}{12pt plus 1fil}
\setlength{\@dblfpbot}{0pt plus 1fil}
\makeatother
\renewcommand{\topfraction}{0.92}
\renewcommand{\bottomfraction}{0.9}
\renewcommand{\textfraction}{0.06}
\renewcommand{\floatpagefraction}{0.72}
\renewcommand{\dbltopfraction}{0.92}
\renewcommand{\dblfloatpagefraction}{0.68}
\setcounter{topnumber}{3}
\setcounter{bottomnumber}{2}
\setcounter{totalnumber}{5}
\setcounter{dbltopnumber}{3}

\section{A. Data Construction and Process Details}
\label{sec:data_details}

\subsection{A.1 RoboTwin Data}
\label{sec:robotwin_data}

\paragraph{Data source and modalities.}
The robot-arm training corpus comprises simulated bimanual manipulation
rollouts from RoboTwin~2.0, generated with the SAPIEN physics engine and the
dual-arm Aloha-AgileX embodiment. Each episode provides synchronized RGB
observations, camera calibration, robot states, end-effector poses, gripper
states, language instructions, and planned action trajectories. We use the
head-camera stream at $1280\times704$ resolution and 10\,fps, obtained by
sampling every third frame from the 30\,fps source video. The synchronized
action at each frame is
\begin{equation}
 [\,q^{L}_{1:6},\ g^L,\ q^{R}_{1:6},\ g^R\,]\in\mathbb R^{14},
 \label{eq:robotwin_action_order}
\end{equation}
where $q^L$ and $q^R$ denote the six joint positions of the left and right
arms, and $g^L$ and $g^R$ denote their gripper coordinates.

\paragraph{Task and visual diversity.}
The training set contains both clean and domain-randomized rollouts. The
randomized regime introduces collision-aware distractors, samples table and
background appearances from more than 12,000 textures, varies illumination,
perturbs the tabletop height by up to 0.03\,m, and displaces the head camera.
Clean backgrounds and extreme illumination each occur in 2\% of randomized
rollouts. The language instructions also vary across equivalent task
executions.

The corpus covers 50 tasks spanning pick-and-place, container placement,
stacking, object ranking, tool use, switch operation, articulated-object
manipulation, bimanual handover, and coordinated dual-arm manipulation. The
task distribution is non-uniform and ranges from 99 to 28,117
training windows per task. The four largest tasks are color-based block ranking
(28,117 windows), size-based block ranking (27,170), bottle placement into a
container (16,881), and block handover (16,723).

\paragraph{Window construction and action normalization.}
Each training example contains 17 consecutive frames with stride one and the
corresponding 17 action vectors. Boundary examples are temporally resampled or
padded to preserve this fixed length. For action dimension $j$, we normalize
using the first and ninety-ninth dataset percentiles, $p_{01,j}$ and $p_{99,j}$:
\begin{equation}
  \widetilde a_j =
  \operatorname{clip}\!\left(
  2\frac{a_j-p_{01,j}}{p_{99,j}-p_{01,j}}-1,\,-1,\,1\right).
  \label{eq:action_normalization}
\end{equation}
RGB frames and actions always share the same temporal indices. Spatial
bucket sampling and matched crop-and-resize augmentation produce the 720p
training inputs.

\subsection{A.2 EgoDex Data}
\label{sec:egodex_data}

\paragraph{Data source and modalities.}
The human-hand training corpus contains approximately 256K egocentric clips
from EgoDex spanning 118 tasks. Each clip is paired with camera intrinsics,
per-frame camera poses, articulated hand transforms and a language instruction. Each RGB clip is paired with a rendered hand-pose video constructed from the capture-time 3D hand tracks.

The tasks cover tabletop setup and cleanup, pick-and-place, stacking,
assembly and disassembly, folding and wrapping, cooking, washing, device
insertion and removal, writing, drawing, and tool use.

\paragraph{Temporal clip construction.}
Each source clip is converted into an 81-frame training clip. Clips longer
than 81 frames are represented by 81 uniformly spaced indices including both
endpoints; clips shorter than 81 frames repeat the final index. The same index sequence is applied to RGB
and hand-pose videos, preserving frame-level correspondence between the target
and condition. Source videos are $1920\times1080$ at 30\,fps, and contiguous
81-frame clips therefore span 2.7 seconds. Training uses matched spatial
transformations at 720p.

\subsection{A.3 Object-Token Grounding}
\label{sec:grounding_details}

\paragraph{Manipulated-object extraction.}
We annotate each training example once with Qwen2.5-0.5B-Instruct. For both
RoboTwin and EgoDex, the task instruction is the sole annotation input; no
video frame, pose signal, temporal metadata, or external visual annotation is
used. The model extracts compact noun phrases for the manipulated objects
while preserving discriminative attributes such as color, material, shape,
label, or container type. Agents, body parts, cameras, supporting surfaces,
backgrounds, and non-interacting scene elements are excluded.

\paragraph{Complete annotation prompt.}
RoboTwin and EgoDex use the same prompt template below, with
\emph{\{instruction\}} replaced by the task instruction for the current
training clip.

\begin{center}
\fbox{\begin{minipage}{0.92\columnwidth}
\small
\textbf{System message}\\
\emph{You are a precise language annotation assistant for manipulation
training. Given only a task instruction, identify the concrete physical
objects directly manipulated by the agent. Return only valid JSON, without
Markdown, explanations, or code fences.}

\medskip
\emph{Use compact English noun phrases and preserve discriminative attributes
stated in the instruction, including color, material, shape, printed label,
object part, and container type. Exclude the robot, hands, fingers, arms,
people, cameras, frames, tables, workspaces, backgrounds, and generic scene
regions unless the instruction explicitly identifies one of them as the
manipulated object. Do not infer objects that are not stated in the
instruction.}

\medskip
\textbf{User message}\\
\emph{Task instruction:}\\
\emph{\{instruction\}}

\medskip
\emph{Return the following JSON object:}\\
\emph{\{}
\hspace*{1em}\emph{``interacting\_objects'': [}\\
\hspace*{2em}\emph{\{``object'': ``compact manipulated-object phrase''\}}\\
\hspace*{1em}\emph{]}\\
\emph{\}}

\medskip
\emph{Include every manipulated object explicitly stated in the instruction.
Preserve the order in which the objects appear. If the instruction refers to
the same object more than once, return it once. Return JSON only.}
\end{minipage}}
\end{center}

\paragraph{Mapping object phrases to text tokens.}
The extracted object phrase is aligned with its occurrence in the original
task instruction after tokenization. All subword tokens associated with the
object phrase are included in the object-token set $\mathcal{O}$. We then
aggregate the cross-attention associated with $\mathcal{O}$ across attention
heads and transformer layers to obtain the object-conditioned attention map
$\mathbf{A}$ used by ADS.

\begin{table*}[t]
\normalsize
\setlength{\tabcolsep}{4pt}
\noindent\resizebox{0.945\textwidth}{!}{%
\begin{tabular}{@{}p{0.11\textwidth}p{0.38\textwidth}p{0.42\textwidth}@{}}
\toprule
Setting & Task instruction & Extracted object phrase \\
\midrule
Robot arm &
``Take the bottle with white printed label from the table and keep upright.'' &
\emph{white printed-label bottle}. \\
\addlinespace
Human hand &
``Pick up the orange plate from the table.'' &
\emph{orange plate}. \\
\bottomrule
\end{tabular}%
}
\captionsetup{justification=raggedright,singlelinecheck=false}
\caption{Representative instruction-only object-grounding examples.}
\label{tab:grounding_examples}
\end{table*}

\subsection{A.4 Hand-Pose Video Construction}
\label{sec:hand_pose_construction}

\paragraph{Pose representation.}
EgoDex provides capture-time 3D hand and finger tracks together with per-frame
camera calibration. We use these annotations directly and do not estimate hand
pose from the compressed RGB video. Each hand is represented by 21
three-dimensional joints, accompanied by a presence mask. The camera metadata
provides a $4\times4$ world-to-camera transformation and four intrinsic
parameters for every frame.

\paragraph{Projection and rendering.}
For frame $t$, a homogeneous world point $\bar{\mathbf p}$ is transformed by
the world-to-camera matrix $\mathbf E_t$ and projected using the intrinsics
$(f_x,f_y,c_x,c_y)_t$:
\begin{equation}
 \mathbf p_c=\mathbf E_t\bar{\mathbf p},\qquad
 u=f_x p_c^x/p_c^z+c_x,\quad
 v=f_y p_c^y/p_c^z+c_y.
\end{equation}
Points with depth $p_c^z\leq0.01$ are excluded. The 21 joints follow the wrist,
thumb, index, middle, ring, and little-finger ordering. Each finger is connected
to the wrist and rendered as an anti-aliased skeleton on a black background.
We draw green edges with width 2, blue joints with radius 3, and fingertips
with radius 5. Pose videos preserve the RGB resolution and are encoded at
30\,fps using H.264 with the YUV420p pixel format and a constant-rate factor of
23.

\paragraph{Temporal and spatial alignment.}
RGB and pose videos contain the same number of decoded frames and use the same
81-frame index sequence for every training segment. The two modalities
therefore remain aligned by frame index throughout temporal sampling. Identical
crop and resize parameters are subsequently applied to both modalities.

\section{B. Implementation Details}
\label{sec:implementation_details}

\subsection{B.1 Robot-Arm Models}
\label{sec:robot_architecture}

\paragraph{Backbone.}
We instantiate the robot-arm setting with two action-conditioned diffusion
transformer backbones: Wan~2.2 TI2V 5B and Cosmos-Predict~2.5 (action). The
Wan~2.2 backbone contains 30 transformer blocks with hidden width $d=3072$, 24
attention heads, and an FFN width of 14,336. It receives a 17-frame RGB
sequence together with the first-frame visual condition and the synchronized
14-DoF action trajectory. The second model is initialized from
Cosmos-Predict2.5-2B and retains its action-conditioning interface for the same
robot-arm setting.

\paragraph{Action encoder.}
Let $\mathbf a\in\mathbb R^{17\times14}$ denote the normalized action
trajectory. We flatten the complete trajectory into 238 scalars and process it
with two independent multilayer perceptrons of identical topology:
\begin{equation}
 \operatorname{MLP}_{q}(\mathbf a)
 =W_{q,2}\,\operatorname{GELU}(W_{q,1}\operatorname{vec}(\mathbf a)+b_{q,1})
 +b_{q,2}
\end{equation}
with hidden width $4d=12{,}288$. The first encoder produces a $d$-dimensional
vector that is added to the timestep embedding. The second produces
$6d=18{,}432$ values, reshaped to $6\times d$, which modulate the six adaptive
normalization components in every transformer block. Learned binary condition
embeddings distinguish action-conditioned and action-dropped examples. Action
information is therefore injected globally through the timestep and adaptive
normalization pathways.

\paragraph{IMPACT configuration.}
For both backbones, \method{} constructs its attention anchor from the
object-token cross-attention produced by the native transformer. ADS samples $K=8$
candidate masks by adding Gaussian perturbations with $\sigma=0.75$ to the
logit of the detached attention anchor on a $5\times8\times12$ coarse grid. The
perturbed maps are trilinearly upsampled to the latent resolution, transformed
with sampling temperature $\tau=1$ and anchor power $\kappa=0.65$, and
thresholded to retain the top $\rho=10\%$ of positions. Detached regional MSEs
score the candidate masks, and centered scores are converted into aggregation
weights with coefficient $\beta=0.7$. IWS uses the resulting soft interaction
map to assign per-position weights in $[1,\gamma]$, with $\gamma=5$.

\subsection{B.2 Human-Hand Model}
\label{sec:human_architecture}

\paragraph{Pose-conditioned input.}
The human-hand model uses the same 5B transformer and conditions on an 81-frame
hand-pose video. RGB targets, the first-frame reference, and pose frames are
encoded by the shared Wan VAE without an additional pose encoder. The VAE
produces 48-channel latents with temporal and spatial compression factors of 4,
16, and 16, yielding 21 latent timesteps for an 81-frame sequence.

The first latent timestep retains the RGB reference condition. At subsequent
timesteps, the 48-channel reference latent is replaced by the temporally aligned
48-channel pose latent. Four binary mask channels are concatenated to form a
52-channel condition $\mathbf y$. Before 3D patch embedding, the model
concatenates the 48-channel noisy video latent $\mathbf x$ with $\mathbf y$,
giving 100 input channels:
\begin{equation}
   \mathbf h_0=\operatorname{PatchEmbed}
   \bigl([\mathbf x;\mathbf y_{\rm mask+pose/ref}]\bigr).
\end{equation}
The action mask applies pose replacement only to conditioned samples. RGB and
pose inputs share the same temporal indices and spatial transformation.

\section{C. Evaluation Details}
\label{sec:evaluation_details}

\subsection{C.1 Evaluation Protocols and Baseline Versions}
\label{sec:baseline_protocol}

WorldArena evaluates 500 held-out episodes from 50 RoboTwin~2.0 tasks. All
decoded submissions have a minimum resolution of $640\times480$ and a frame
rate of 24\,fps. Text-conditioned submissions contain 121 frames.
Action-conditioned submissions follow the benchmark action sequence and match
the corresponding reference trajectory length. EgoDex models are evaluated at
their release-specific inference settings, as detailed in
\tabref{tab:baseline_specs}.

\begin{center}
\begin{minipage}{\columnwidth}
\centering
\small
\setlength{\tabcolsep}{3pt}
\begin{tabular*}{\columnwidth}{@{\extracolsep{\fill}}lccc@{}}
\toprule
Method & Version & Resolution & Sec. / FPS \\
\midrule
\rowcolor{gray!15}\multicolumn{4}{l}{\emph{General video generation}} \\
HunyuanVideo-1.5 & 25.11.20 & $848\times480$ & 5 / 24 \\
Cosmos-Predict 2.5 & 25.10.06 & $1280\times704$ & 5 / 16 \\
\midrule
\rowcolor{gray!15}\multicolumn{4}{l}{\emph{Pose-controlled video generation}} \\
MimicMotion & 24.07.08 & $1024\times576$ & 4.8 / 15 \\
MagicPose & 24.04.03 & $512\times512$ & 4 / 15 \\
VACE & 25.03.11 & $720\times1080$ & 5 / 16 \\
LOME & 26.04.05 & $832\times480$ & 5 / 15 \\
\midrule
\rowcolor{gray!15}\multicolumn{4}{l}{\emph{Wan 2.2-based models}} \\
Wan 2.2 & 25.07.28 & $1280\times704$ & 5 / 24 \\
Wan 2.2-AC / +\method{} & -- & $1280\times704$ & 3.4 / 24 \\
\bottomrule
\end{tabular*}
\captionof{table}{Detailed inference configurations of video generation models
evaluated on EgoDex. All models use image-to-video generation; ``--'' denotes
our adapted variants.}
\label{tab:baseline_specs}
\end{minipage}
\end{center}

Metric computation uses the ordered decoded frames from each output. For
Wan~2.2-AC and \method{}, the 81 output frames and the pose condition use
identical temporal indices. Each model uses its official sampling schedule and
guidance configuration.

\subsection{C.2 WorldArena Metric Definitions}
\label{sec:worldarena_metrics}

WorldArena normalizes each raw metric to $[0,1]$ using empirically selected
boundaries and reports EWMScore as $100$ times the arithmetic mean of the 16
normalized metrics.  Thus, all displayed entries are higher-is-better, including
metrics whose underlying raw quantity is an error.

\begin{table*}[p]
\centering
\normalsize
\setlength{\tabcolsep}{5pt}
\renewcommand{\arraystretch}{1.80}
\begin{tabular}{@{}p{0.10\textwidth}p{0.24\textwidth}p{0.57\textwidth}@{}}
\toprule
Dimension & Metric & implementation \\
\midrule
\multirow{3}{*}{Visual}
& IQ (Image Quality) & Frame clarity and distortion quality from the no-reference MUSIQ image-quality model; averaged over frames. \\
& AQ (Aesthetic Quality) & Per-frame visual appeal (lighting, color, composition) from the LAION aesthetic predictor; averaged over frames. \\
& JS (JEPA Similarity) & Distributional similarity between generated and ground-truth V-JEPA video features, computed using MMD with a second-order polynomial kernel. \\
\midrule
\multirow{3}{*}{Motion}
& DD (Dynamic Degree) & Salient motion intensity: RAFT flow between adjacent frames, averaging the top 5\% motion magnitudes and applying a resolution-adaptive sigmoid. \\
& FS (Flow Score) & Overall motion intensity: mean RAFT optical-flow magnitude over all pixels and adjacent-frame pairs. \\
& MS (Motion Smoothness) & Temporal smoothness: SSIM between each actual middle frame and a VFI-Mamba interpolation from its neighbors, weighted by log motion magnitude to avoid rewarding static video. \\
\midrule
\multirow{3}{*}{Content}
& SC (Subject Consistency) & Subject identity/structure stability from DINO cosine similarity of each frame to both the first and previous frame, penalized for near-static video. \\
& BC (Background Consistency) & Global background/scene stability from CLIP image-feature cosine similarity to the first and previous frames, with the same low-motion penalty. \\
& PC (Photometric Consistency) & Pixel-level texture stability from forward--backward SEA-RAFT flow cycle error (raw AEPE is lower-better; the reported consistency score is inverted/normalized). \\
\midrule
\multirow{2}{*}{Physics}
& Inter. (Interaction Quality) & Qwen3-VL 1--5 judgment of physically plausible robot--object contact, force transmission, and interaction, divided by five. \\
& Traj. (Trajectory Accuracy) & Robot-arm trajectory agreement with ground truth: SAM~3 arm boxes/trajectories compared through normalized dynamic time warping. \\
\midrule
\multirow{2}{*}{3D}
& Depth (Depth Accuracy) & Monocular depth agreement to ground truth after per-video median-scale alignment; based on depth error and converted to a higher-is-better accuracy. Up to 40 frames are sampled uniformly. \\
& Persp. (Perspectivity) & Qwen3-VL judgment of 3D plausibility, including scale-versus-depth, lighting, and occlusion relations. \\
\midrule
\multirow{3}{*}{Control}
& Instr. (Instruction Following) & Qwen3-VL judgment of whether action type, target object, and resulting task state follow the instruction. \\
& Sem. (Semantic Alignment) & Cosine similarity between Qwen2.5-VL descriptions of generated and reference videos. \\
& Act. (Action Following) & Response diversity under three distinct instructions sharing one initial frame; average pairwise feature dissimilarity between the three generated videos. \\
\bottomrule
\end{tabular}
\caption{Meaning and implementation of all 16 WorldArena metrics.}
\label{tab:worldarena_metric_definitions}
\end{table*}

\paragraph{Full 16-metric results.}
\label{sec:full_worldarena_results}

Tables~\ref{tab:worldarena_full_quality} and~\ref{tab:worldarena_full_task}
report all 16 normalized WorldArena metrics for the same models and ordering.
The first covers the generation-quality dimensions (visual quality, motion
quality, content consistency) and the overall EWMScore; the second covers the
task-oriented dimensions (physics adherence, 3D accuracy, controllability).

\begin{table*}[p]
\centering
\small
\setlength{\tabcolsep}{4.2pt}
\renewcommand{\arraystretch}{0.90}
\begin{tabular}{@{}l c ccc ccc ccc@{}}
\toprule
\multirow{2}{*}{Model} & \multirow{2}{*}{EWMScore}
& \multicolumn{3}{c}{Visual Quality}
& \multicolumn{3}{c}{Motion Quality}
& \multicolumn{3}{c}{Content Consistency} \\
\cmidrule(lr){3-5}\cmidrule(lr){6-8}\cmidrule(lr){9-11}
& & IQ & AQ & JS & DD & FS & MS & SC & BC & PC \\
\midrule
\rowcolor{gray!15}\multicolumn{11}{l}{\emph{General world models}} \\
CogVideoX & 57.90 & 0.3582 & 0.3777 & \textbf{0.9384} & 0.3166 & 0.2189 & 0.7391 & 0.8083 & 0.8773 & \textbf{0.3580} \\
Veo 3.1 & 58.87 & 0.6605 & \textbf{0.4632} & 0.5694 & 0.5450 & 0.1396 & 0.6989 & 0.7878 & 0.8710 & 0.3247 \\
Wan 2.6 & 61.86 & \textbf{0.6824} & 0.4433 & 0.7229 & \textbf{0.7421} & \underline{0.4532} & \underline{0.8539} & 0.7517 & 0.8687 & 0.1904 \\
\midrule
\rowcolor{gray!15}\multicolumn{11}{l}{\emph{Embodied world models}} \\
GigaWorld-0 & 53.39 & 0.5041 & 0.3991 & 0.4413 & 0.6709 & 0.3118 & 0.7811 & 0.7303 & 0.8563 & 0.1756 \\
Genie Envisioner & 43.65 & 0.2305 & 0.3289 & 0.3340 & \underline{0.6930} & 0.0855 & 0.6966 & 0.7760 & 0.9024 & 0.2006 \\
Vidar & 51.60 & 0.4145 & 0.4068 & 0.5608 & 0.2767 & 0.1426 & 0.7973 & 0.7629 & 0.8300 & 0.2350 \\
IRASim & 58.12 & 0.3489 & 0.3623 & 0.9330 & 0.4139 & 0.2083 & 0.7052 & \underline{0.8312} & 0.9068 & 0.3522 \\
CtrlWorld & 59.70 & 0.3522 & 0.3893 & 0.9185 & 0.4257 & 0.3449 & 0.7377 & \textbf{0.8411} & 0.9057 & 0.1729 \\
\midrule
\rowcolor{gray!15}\multicolumn{11}{l}{\emph{Representation-guided models}} \\
TesserAct & 53.23 & 0.3322 & \underline{0.4590} & 0.4579 & 0.5150 & 0.2447 & 0.7579 & 0.8250 & \textbf{0.9238} & 0.2491 \\
RoboMaster & 51.84 & 0.3487 & 0.3842 & 0.2966 & 0.6124 & 0.1484 & 0.6940 & 0.8295 & \underline{0.9123} & 0.3356 \\
WoW & 54.88 & 0.4587 & 0.3868 & 0.7440 & 0.4608 & 0.2706 & 0.7692 & 0.8161 & 0.9025 & 0.2170 \\
\midrule
\rowcolor{gray!15}\multicolumn{11}{l}{\emph{Wan 2.2-based models}} \\
Wan 2.2 & 50.79 & 0.3884 & 0.3963 & 0.7575 & 0.4349 & 0.1269 & 0.7019 & 0.7400 & 0.8000 & 0.0806 \\
Wan 2.2-AC & 58.65 & 0.4053 & 0.3575 & \underline{0.9343} & 0.4262 & 0.2488 & 0.7719 & 0.8250 & 0.9095 & 0.0821 \\
Wan 2.2-AC + \method{} (w/o ADS) & 61.54 & 0.5236 & 0.4078 & 0.8819 & 0.4381 & 0.2611 & 0.7756 & 0.8233 & 0.8970 & 0.1090 \\
Wan 2.2-AC + \method{} & \underline{62.46} & 0.4906 & 0.4223 & 0.9050 & 0.4630 & 0.3328 & 0.8026 & 0.8191 & 0.8953 & 0.0944 \\
\midrule
\rowcolor{gray!15}\multicolumn{11}{l}{\emph{Cosmos-based models}} \\
Cosmos-Predict 2.5 (text) & 50.81 & \underline{0.6668} & 0.4501 & 0.3126 & 0.5911 & 0.4302 & 0.7882 & 0.7488 & 0.8511 & 0.1383 \\
Cosmos-Predict 2.5 (action) & 55.91 & 0.4489 & 0.3576 & 0.9296 & 0.3994 & 0.0573 & 0.7100 & 0.8197 & 0.8894 & \underline{0.3528} \\
\quad + \method{} & \textbf{62.53} & 0.5588 & 0.3941 & 0.7366 & 0.5810 & \textbf{0.5816} & \textbf{0.8839} & 0.8026 & 0.8862 & 0.0482 \\
\bottomrule
\end{tabular}
\caption{Full WorldArena results (1/2): EWMScore and the nine
generation-quality metrics. EWMScore is the mean of all 16 metrics on a $[0,100]$ scale;
components are normalized to $[0,1]$. Higher is better ($\uparrow$); boldface
and underlining mark the best and second-best per column.}
\label{tab:worldarena_full_quality}

\vspace{6pt}

\small
\setlength{\tabcolsep}{6pt}
\renewcommand{\arraystretch}{0.90}
\resizebox{0.945\textwidth}{!}{%
\begin{tabular}{@{}l cc cc ccc@{}}
\toprule
\multirow{2}{*}{Model}
& \multicolumn{2}{c}{Physics Adherence}
& \multicolumn{2}{c}{3D Accuracy}
& \multicolumn{3}{c}{Controllability} \\
\cmidrule(lr){2-3}\cmidrule(lr){4-5}\cmidrule(lr){6-8}
& Inter. & Traj. & Depth & Persp. & Instr. & Sem. & Act. \\
\midrule
\rowcolor{gray!15}\multicolumn{8}{l}{\emph{General world models}} \\
CogVideoX & 0.5940 & 0.3526 & 0.9097 & 0.7828 & 0.7268 & \textbf{0.8977} & 0.0076 \\
Veo 3.1 & \textbf{0.7872} & 0.1231 & 0.7421 & 0.8276 & \textbf{0.9328} & 0.8607 & 0.0852 \\
Wan 2.6 & 0.7280 & 0.1182 & 0.7144 & 0.8032 & 0.8536 & 0.8728 & 0.0992 \\
\midrule
\rowcolor{gray!15}\multicolumn{8}{l}{\emph{Embodied world models}} \\
GigaWorld-0 & 0.5368 & 0.1552 & 0.6316 & 0.7596 & 0.6156 & 0.8591 & 0.1134 \\
Genie Envisioner & 0.2052 & 0.0679 & 0.8663 & 0.5284 & 0.2028 & 0.8544 & 0.0109 \\
Vidar & 0.5348 & 0.1928 & 0.7872 & 0.7592 & 0.5912 & 0.8826 & 0.0819 \\
IRASim & 0.5656 & 0.3639 & \textbf{0.9312} & 0.7788 & 0.6604 & 0.8849 & 0.0526 \\
CtrlWorld & 0.6212 & \textbf{0.4766} & \underline{0.9300} & 0.7960 & 0.7272 & \underline{0.8912} & 0.0210 \\
\midrule
\rowcolor{gray!15}\multicolumn{8}{l}{\emph{Representation-guided models}} \\
TesserAct & 0.5800 & 0.1396 & 0.7159 & 0.7920 & 0.6152 & 0.8783 & 0.0311 \\
RoboMaster & 0.5364 & 0.1158 & 0.8335 & 0.7588 & 0.5772 & 0.8761 & 0.0352 \\
WoW & 0.5564 & 0.2058 & 0.7283 & 0.7672 & 0.5692 & 0.8842 & 0.0434 \\
\midrule
\rowcolor{gray!15}\multicolumn{8}{l}{\emph{Wan 2.2-based models}} \\
Wan 2.2 & 0.5184 & 0.1627 & 0.7768 & 0.7660 & 0.5376 & 0.8877 & 0.0512 \\
Wan 2.2-AC & 0.6672 & \underline{0.4009} & 0.8657 & 0.8574 & 0.7394 & 0.8837 & 0.0089 \\
Wan 2.2-AC + \method{} (w/o ADS) & \underline{0.7546} & 0.3485 & 0.8974 & \textbf{0.9780} & 0.8472 & 0.8886 & 0.0143 \\
Wan 2.2-AC + \method{} & 0.7516 & 0.3657 & 0.9026 & \underline{0.9486} & \underline{0.8560} & 0.8881 & 0.0559 \\
\midrule
\rowcolor{gray!15}\multicolumn{8}{l}{\emph{Cosmos-based models}} \\
Cosmos-Predict 2.5 (text) & 0.3872 & 0.0816 & 0.7051 & 0.7964 & 0.2664 & 0.7733 & \underline{0.1418} \\
Cosmos-Predict 2.5 (action) & 0.5500 & 0.2945 & 0.8862 & 0.7644 & 0.5840 & 0.8879 & 0.0133 \\
\quad + \method{} & 0.6360 & 0.2003 & 0.6557 & 0.9040 & 0.6360 & 0.8738 & \textbf{0.6260} \\
\bottomrule
\end{tabular}%
}
\caption{Full WorldArena results (2/2): the task-oriented metrics.}
\label{tab:worldarena_full_task}
\end{table*}

\clearpage
\onecolumn
\captionsetup[figure]{font=normalsize,skip=6pt}
\newcommand{\appendixcaseimg}[1]{%
  \includegraphics[width=\textwidth,height=0.318\textheight,keepaspectratio]{#1}%
}
\newcommand{\appendixcaseblock}[4]{%
  \begin{center}
    \appendixcaseimg{#1}\\[0.25em]
    {\normalsize\textbf{Prompt:} #2\par}
    \captionof{figure}{#3}
    \label{#4}
  \end{center}
}
\makeatletter
\setlength{\@fptop}{2pt}
\setlength{\@fpsep}{8pt}
\setlength{\@fpbot}{0pt plus 1fil}
\makeatother
\renewcommand{\floatpagefraction}{0.50}
\subsection{C.3 Additional Robot-Arm Cases}
\label{sec:additional_robot_cases}
In this section, we present more qualitative results of robot-arm manipulation
on WorldArena.

\appendixcaseblock
  {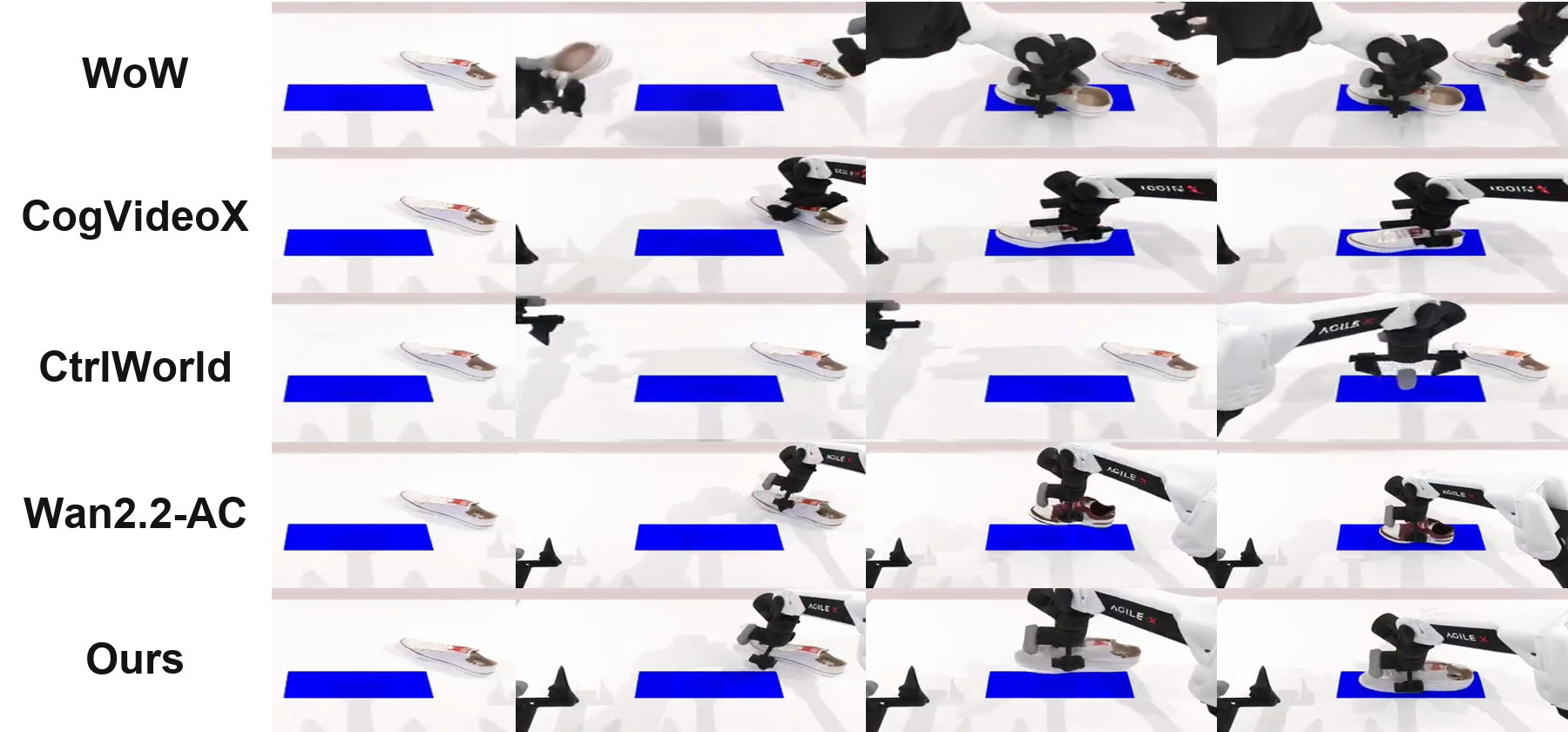}
  {Pick up the \textcolor{red}{\textbf{printed sneaker}} and place it on the blue mat.}
  {Printed-sneaker placement comparison. Rows show WoW, CogVideoX,
  CtrlWorld, Wan~2.2-AC, and \method{} (Ours) under the same initial observation,
  instruction, and action trajectory; columns are uniformly sampled frames from
  each rollout.}
  {fig:robot_case_0038}

\appendixcaseblock
  {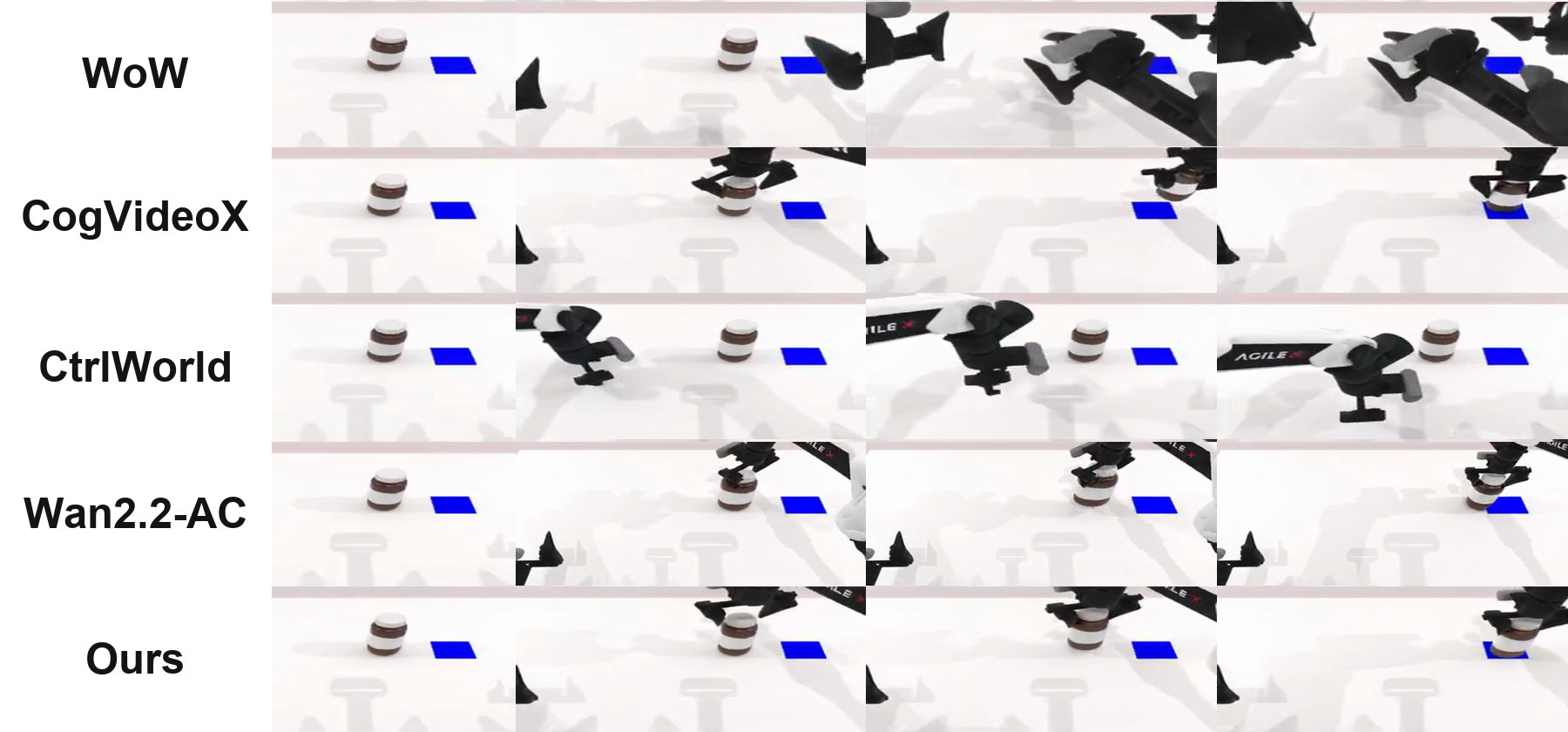}
  {Pick up the \textcolor{red}{\textbf{brown-and-white bottle}} and place it on the blue mat.}
  {Bottle placement comparison. Rows show WoW, CogVideoX, CtrlWorld,
  Wan~2.2-AC, and \method{} (Ours) under the same initial observation,
  instruction, and action trajectory; columns are uniformly sampled frames from
  each rollout.}
  {fig:robot_case_0042}
\clearpage

\begin{figure}[p]
\centering
\appendixcaseimg{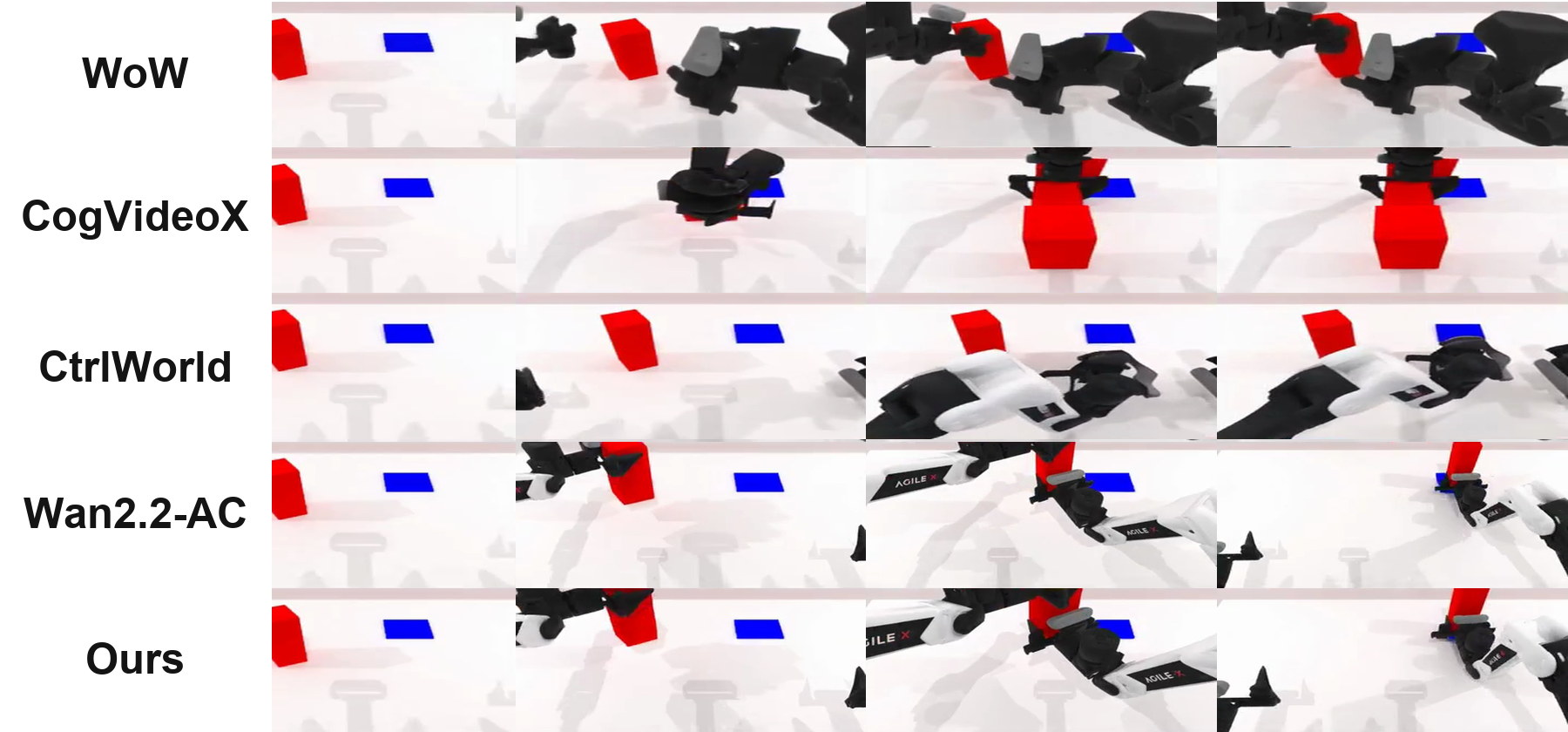}
\par\smallskip
{\normalsize\textbf{Prompt:} Pick up the \textcolor{red}{\textbf{red block}} and
place it on the blue target.\par}
\caption{Block placement comparison. Rows show WoW, CogVideoX, CtrlWorld,
Wan~2.2-AC, and \method{} (Ours) under the same initial observation,
instruction, and action trajectory; columns are uniformly sampled frames from
each rollout.}
\label{fig:robot_case_0047}
\end{figure}

\begin{figure}[p]
\centering
\appendixcaseimg{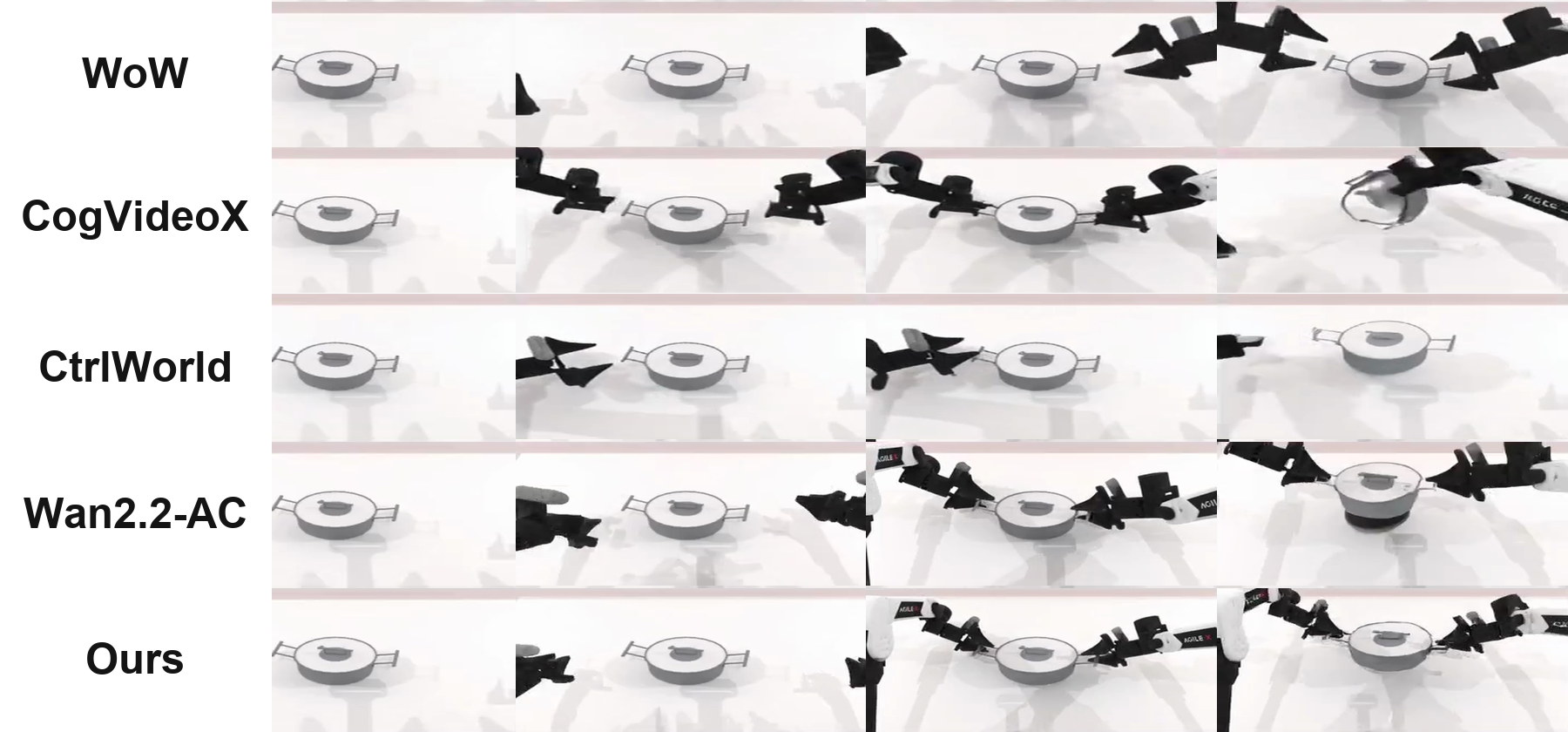}
\par\smallskip
{\normalsize\textbf{Prompt:} Grasp the \textcolor{red}{\textbf{lidded pot}} with
both grippers and lift it.\par}
\caption{Bimanual pot-lifting comparison. Rows show WoW, CogVideoX,
CtrlWorld, Wan~2.2-AC, and \method{} (Ours) under the same initial observation,
instruction, and action trajectory; columns are uniformly sampled frames from
each rollout.}
\label{fig:robot_case_0059}
\end{figure}
\clearpage

\begin{figure}[p]
\centering
\appendixcaseimg{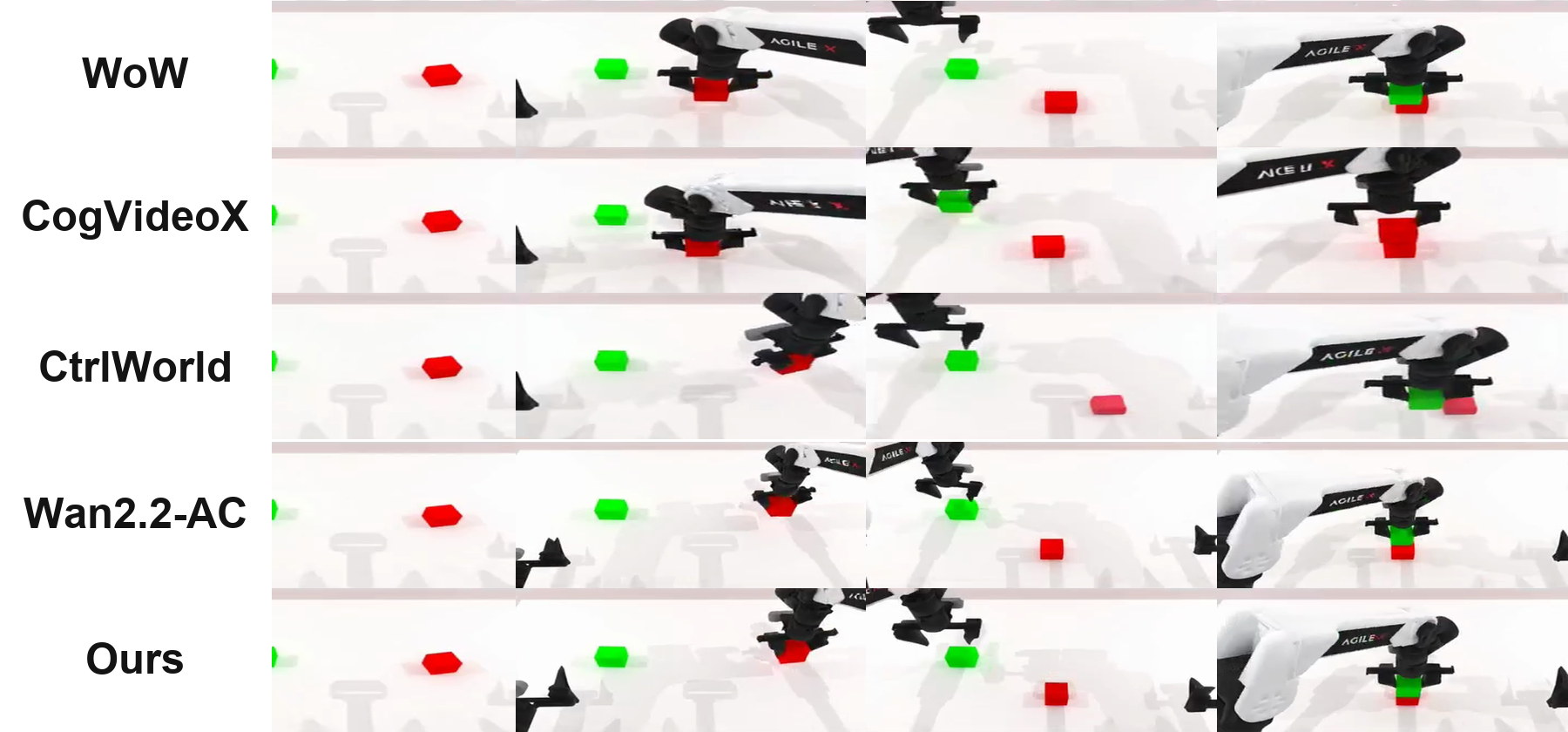}
\par\smallskip
{\normalsize\textbf{Prompt:} Pick up the \textcolor{red}{\textbf{green block}} and
stack it on the red block.\par}
\caption{Block-stacking comparison. Rows show WoW, CogVideoX, CtrlWorld,
Wan~2.2-AC, and \method{} (Ours) under the same initial observation,
instruction, and action trajectory; columns are uniformly sampled frames from
each rollout.}
\label{fig:robot_case_0080}
\end{figure}

\begin{figure}[p]
\centering
\appendixcaseimg{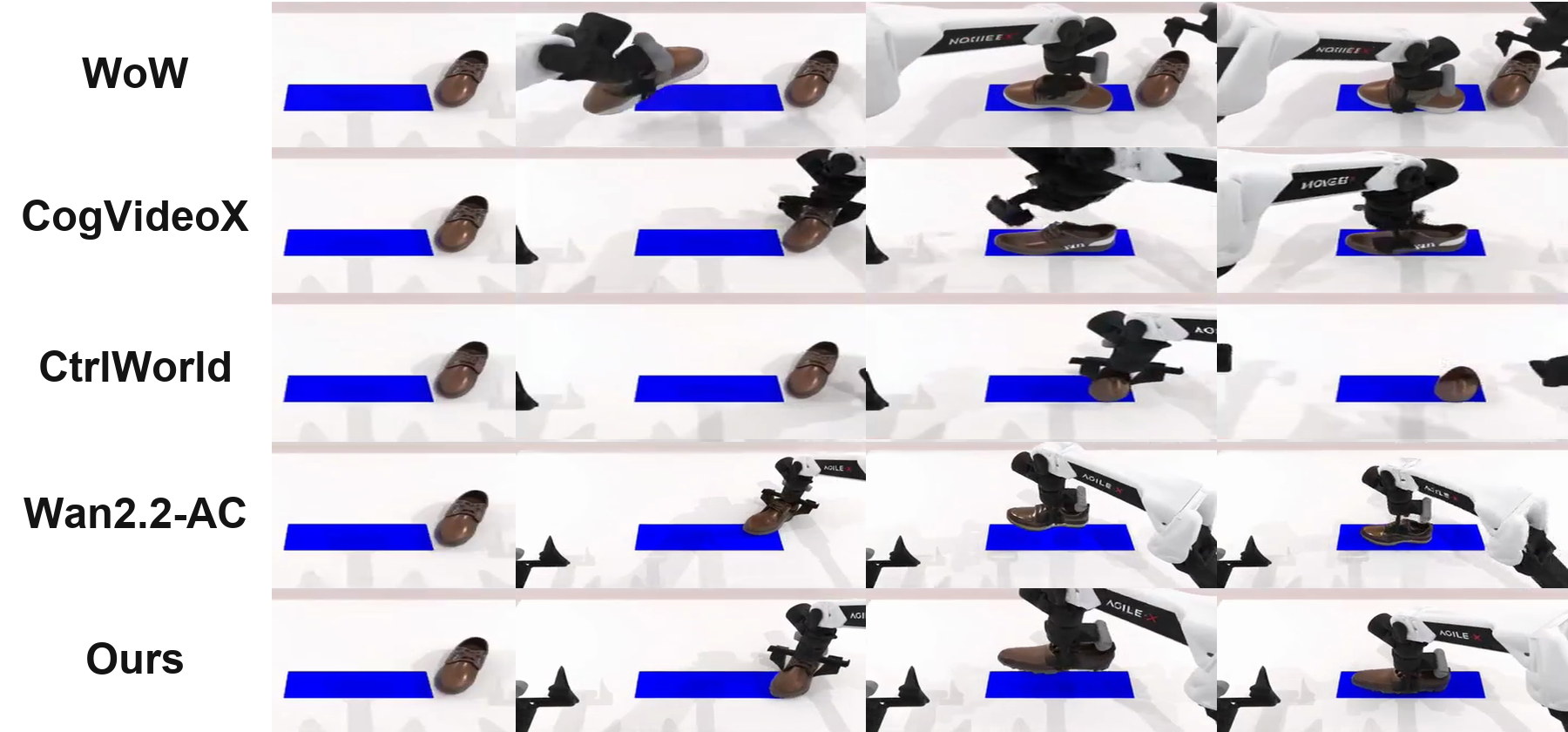}
\par\smallskip
{\normalsize\textbf{Prompt:} Pick up the \textcolor{red}{\textbf{brown shoe}} and
place it on the blue mat.\par}
\caption{Shoe placement comparison. Rows show WoW, CogVideoX, CtrlWorld,
Wan~2.2-AC, and \method{} (Ours) under the same initial observation,
instruction, and action trajectory; columns are uniformly sampled frames from
each rollout.}
\label{fig:robot_case_0108}
\end{figure}
\clearpage

\begin{figure}[p]
\centering
\appendixcaseimg{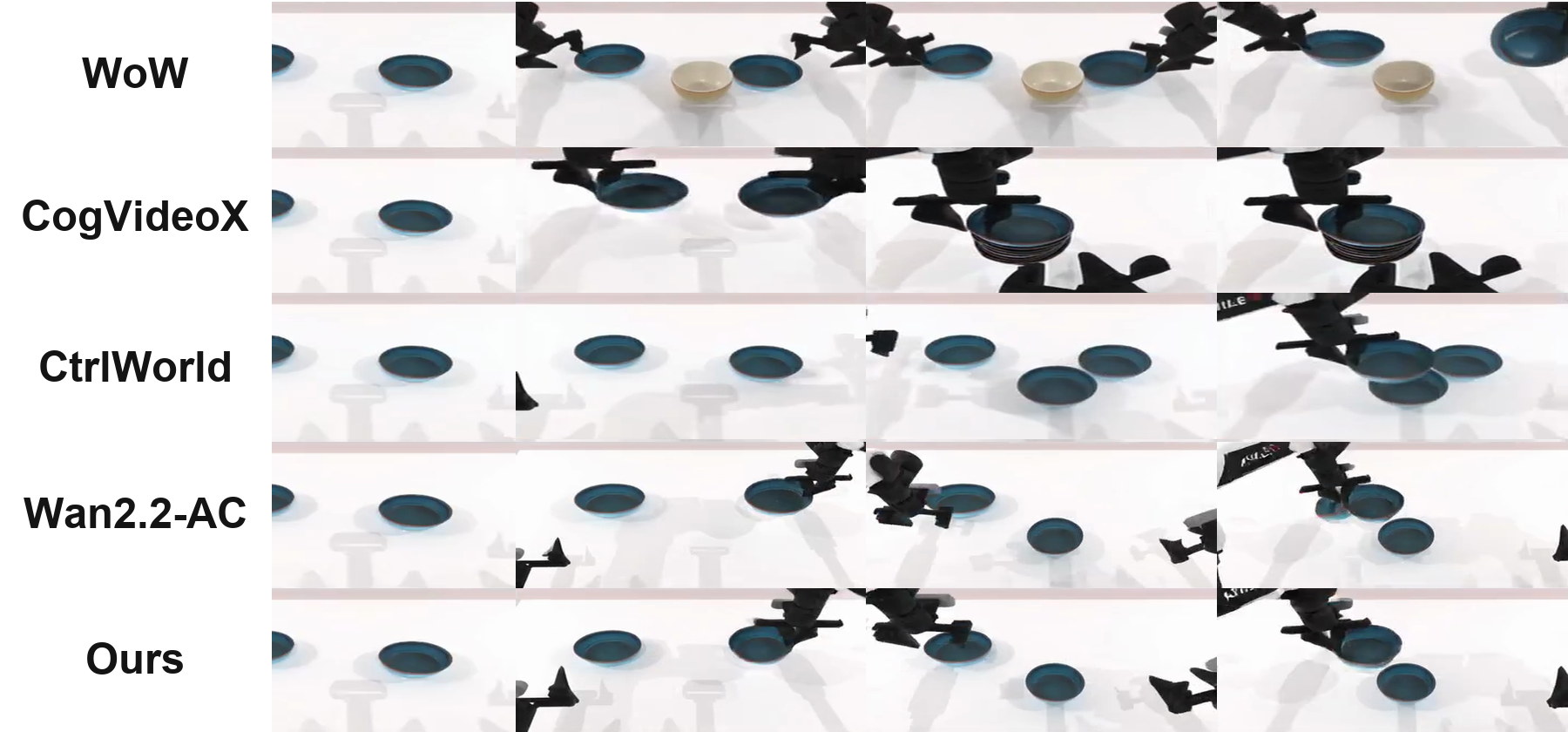}
\par\smallskip
{\normalsize\textbf{Prompt:} Pick up one \textcolor{red}{\textbf{blue bowl}} and
stack it inside the other blue bowl.\par}
\caption{Bowl-stacking comparison. Rows show WoW, CogVideoX, CtrlWorld,
Wan~2.2-AC, and \method{} (Ours) under the same initial observation,
instruction, and action trajectory; columns are uniformly sampled frames from
each rollout.}
\label{fig:robot_case_0143}
\end{figure}

\begin{figure}[p]
\centering
\appendixcaseimg{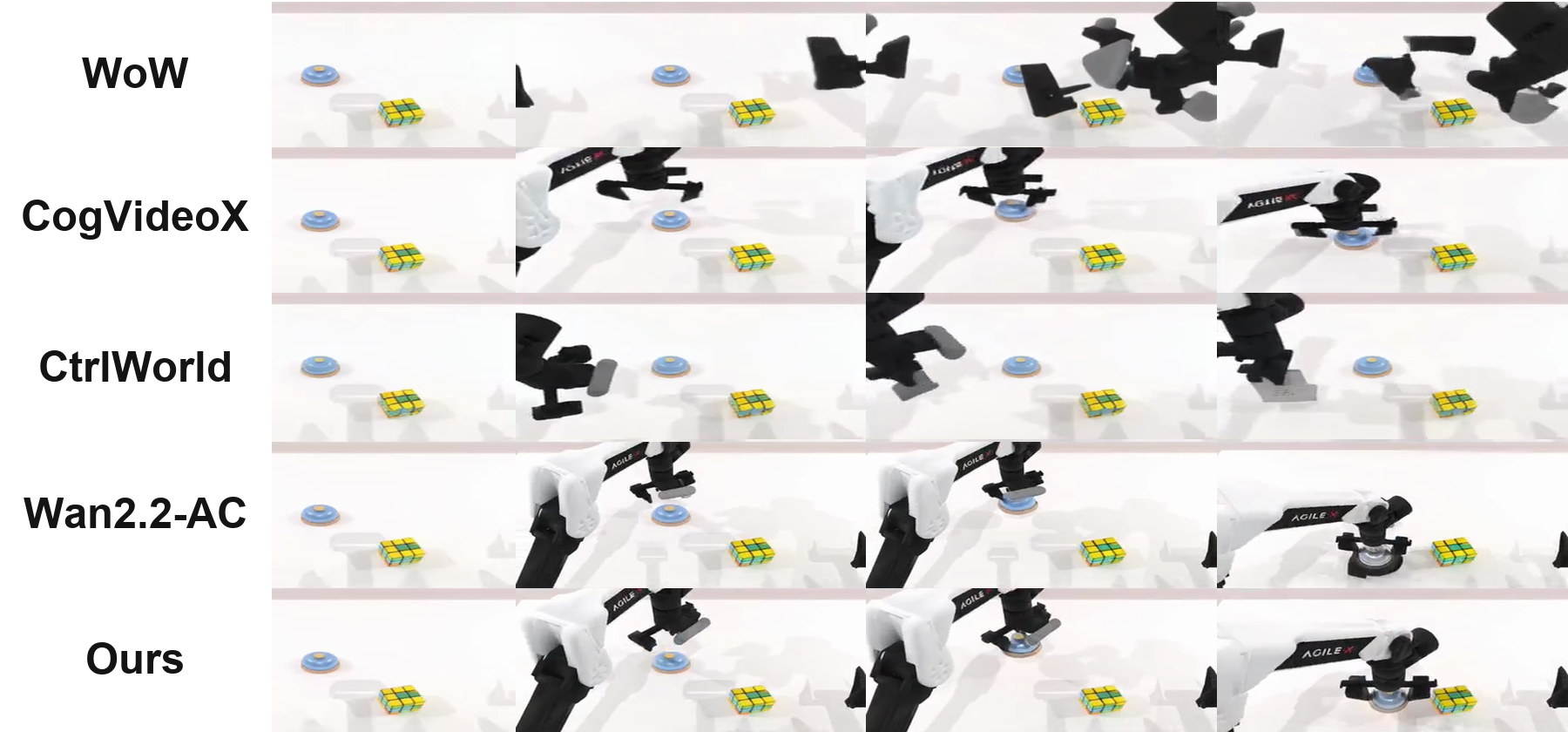}
\par\smallskip
{\normalsize\textbf{Prompt:} Press the
\textcolor{red}{\textbf{blue service bell}}.\par}
\caption{Service-bell pressing comparison. Rows show WoW, CogVideoX,
CtrlWorld, Wan~2.2-AC, and \method{} (Ours) under the same initial observation,
instruction, and action trajectory; columns are uniformly sampled frames from
each rollout.}
\label{fig:robot_case_0169}
\end{figure}
\clearpage

\begin{figure}[p]
\centering
\appendixcaseimg{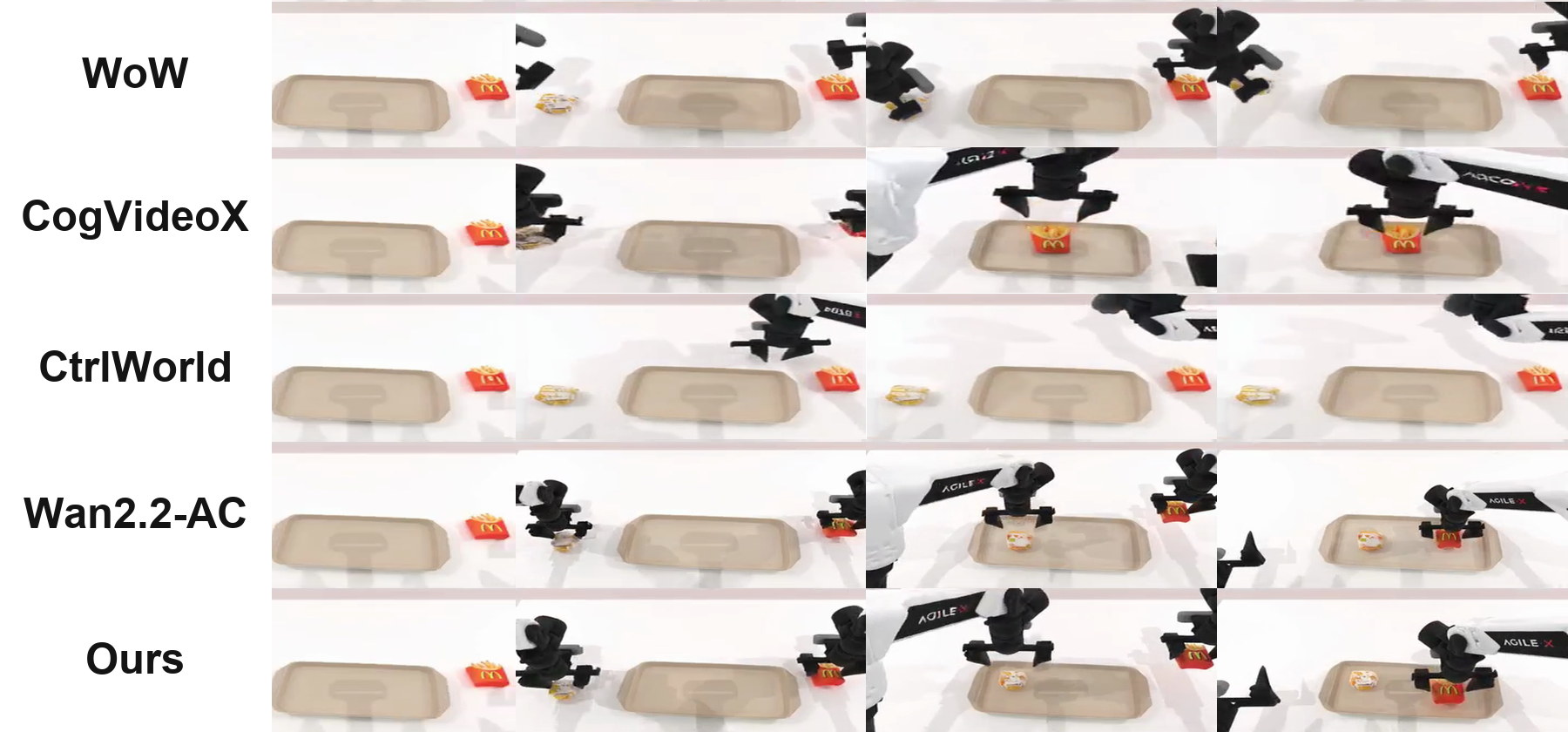}
\par\smallskip
{\normalsize\textbf{Prompt:} Place the
\textcolor{red}{\textbf{toy hamburger and French fries}} in the tray.\par}
\caption{Food-toy placement comparison. Rows show WoW, CogVideoX, CtrlWorld,
Wan~2.2-AC, and \method{} (Ours) under the same initial observation,
instruction, and action trajectory; columns are uniformly sampled frames from
each rollout.}
\label{fig:robot_case_0220}
\end{figure}

\begin{figure}[p]
\centering
\appendixcaseimg{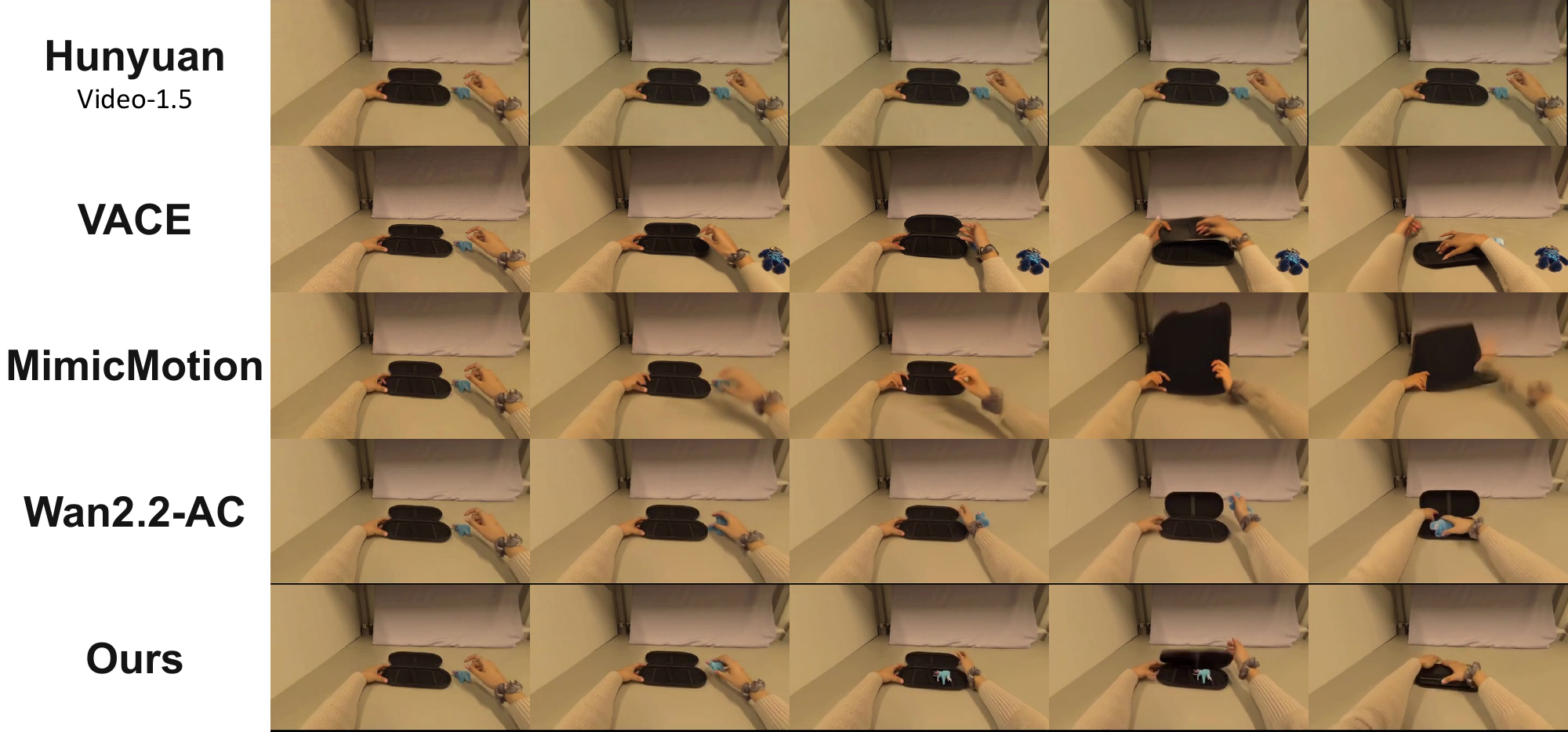}
\par\smallskip
{\normalsize\textbf{Prompt:} Pick up the
\textcolor{red}{\textbf{blue elephant toy}} from beside the black case.\par}
\caption{Elephant-toy pickup comparison. Rows show HunyuanVideo-1.5, VACE,
MimicMotion, Wan~2.2-AC, and \method{} (Ours) under the same initial observation
and task instruction; columns are uniformly sampled frames from each rollout.}
\label{fig:hand_case_01}
\end{figure}
\clearpage
\subsection{C.4 Additional Human-Hand Cases}
\label{sec:additional_hand_cases}
In this section, we present more qualitative results of human-hand manipulation
on EgoDex.

\appendixcaseblock
  {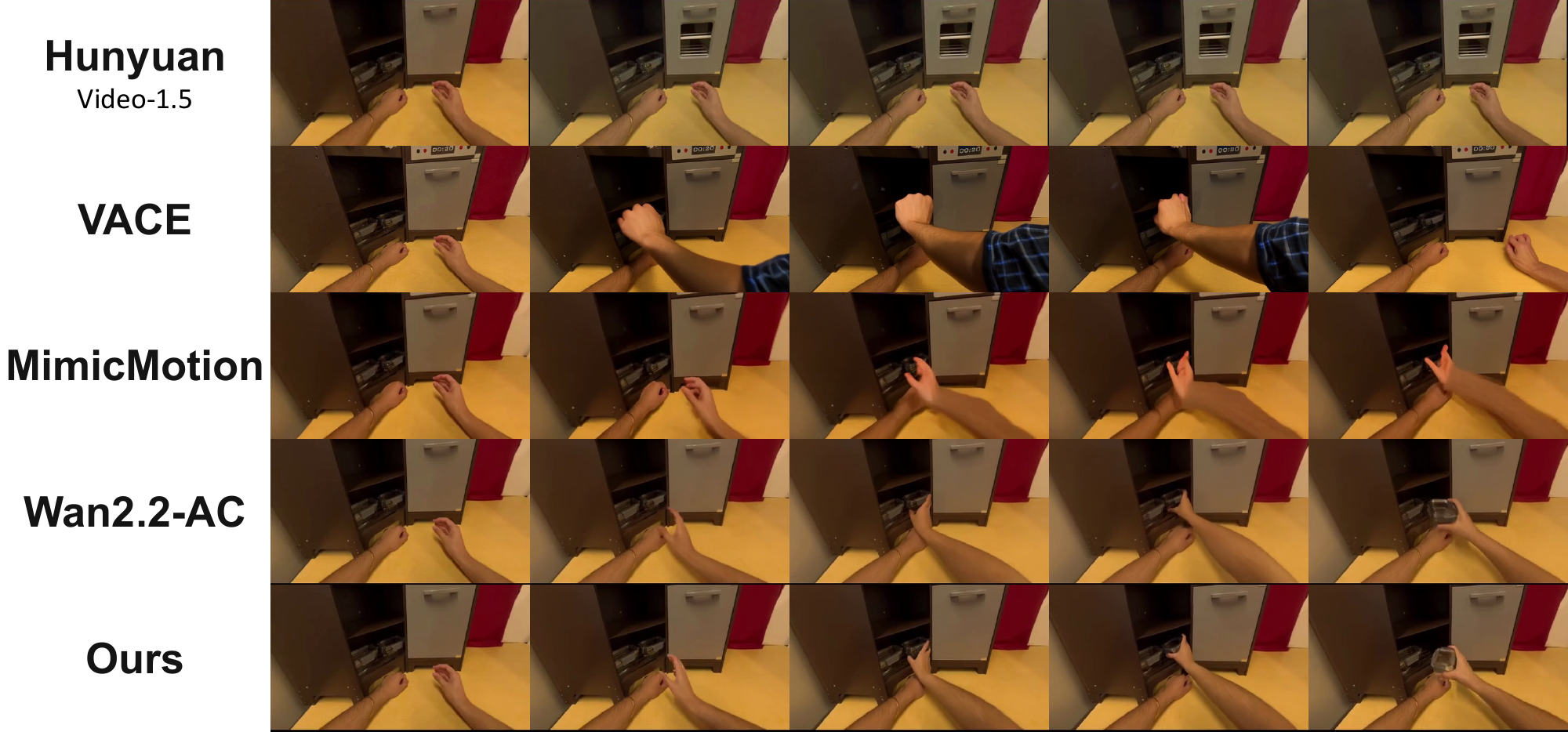}
  {Grasp the \textcolor{red}{\textbf{transparent container}} and lift it from the shelf.}
  {Container-lifting comparison. Rows show HunyuanVideo-1.5, VACE,
  MimicMotion, Wan~2.2-AC, and \method{} (Ours) under the same initial observation
  and task instruction; columns are uniformly sampled frames from each rollout.}
  {fig:hand_case_02}

\appendixcaseblock
  {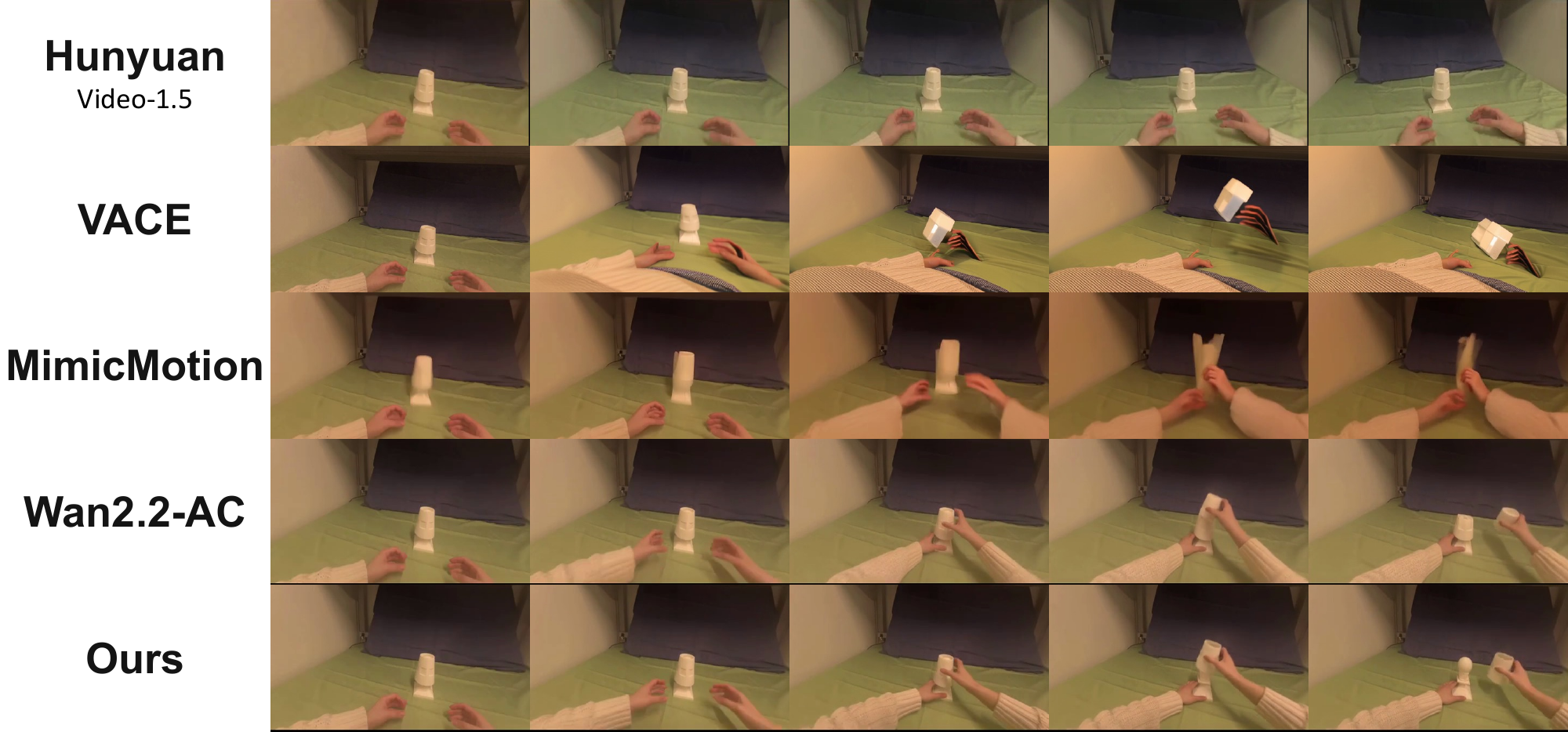}
  {Grasp the \textcolor{red}{\textbf{white cylindrical object}} and lift it from the base.}
  {Cylindrical-object lifting comparison. Rows show HunyuanVideo-1.5,
  VACE, MimicMotion, Wan~2.2-AC, and \method{} (Ours) under the same initial
  observation and task instruction; columns are uniformly sampled frames from
  each rollout.}
  {fig:hand_case_03}
\clearpage

\begin{figure}[p]
\centering
\appendixcaseimg{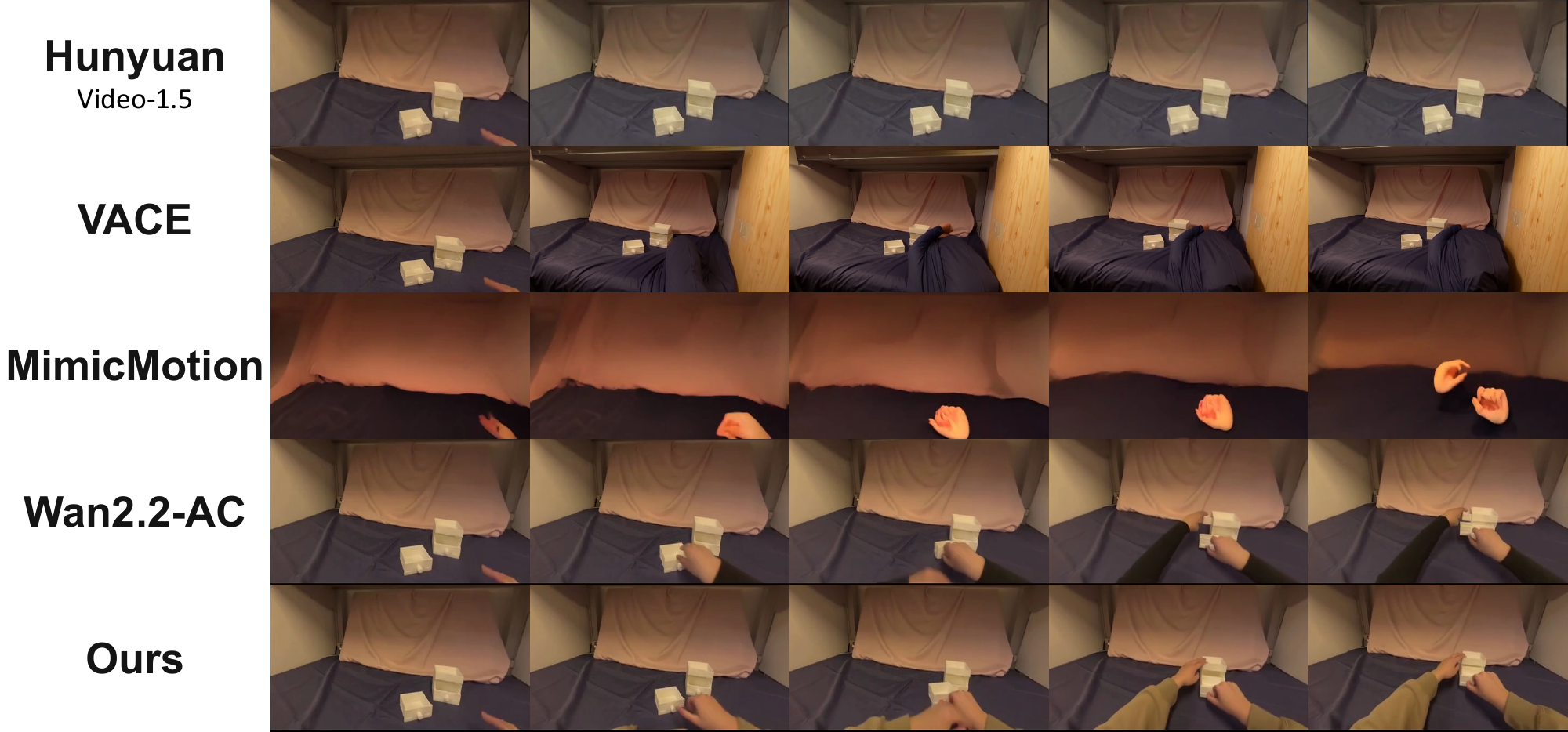}
\par\smallskip
{\normalsize\textbf{Prompt:} Pick up the
\textcolor{red}{\textbf{two white drawers}} and stack them together.\par}
\caption{Drawer-stacking comparison. Rows show HunyuanVideo-1.5, VACE,
MimicMotion, Wan~2.2-AC, and \method{} (Ours) under the same initial observation
and task instruction; columns are uniformly sampled frames from each rollout.}
\label{fig:hand_case_04}
\end{figure}

\begin{figure}[p]
\centering
\appendixcaseimg{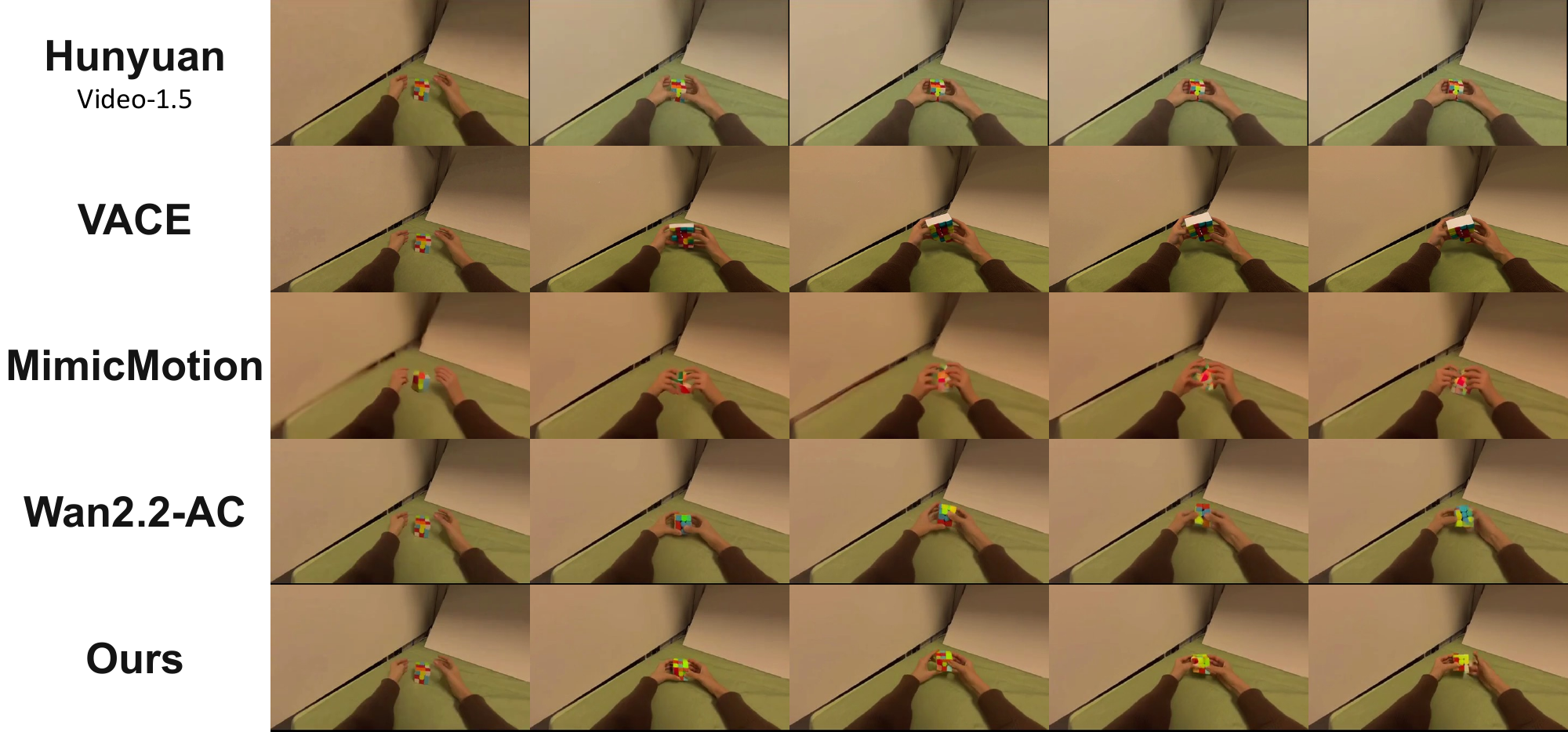}
\par\smallskip
{\normalsize\textbf{Prompt:} Grasp and adjust the
\textcolor{red}{\textbf{colorful block structure}}.\par}
\caption{Block-structure adjustment comparison. Rows show HunyuanVideo-1.5,
VACE, MimicMotion, Wan~2.2-AC, and \method{} (Ours) under the same initial
observation and task instruction; columns are uniformly sampled frames from
each rollout.}
\label{fig:hand_case_05}
\end{figure}
\clearpage

\begin{figure}[p]
\centering
\appendixcaseimg{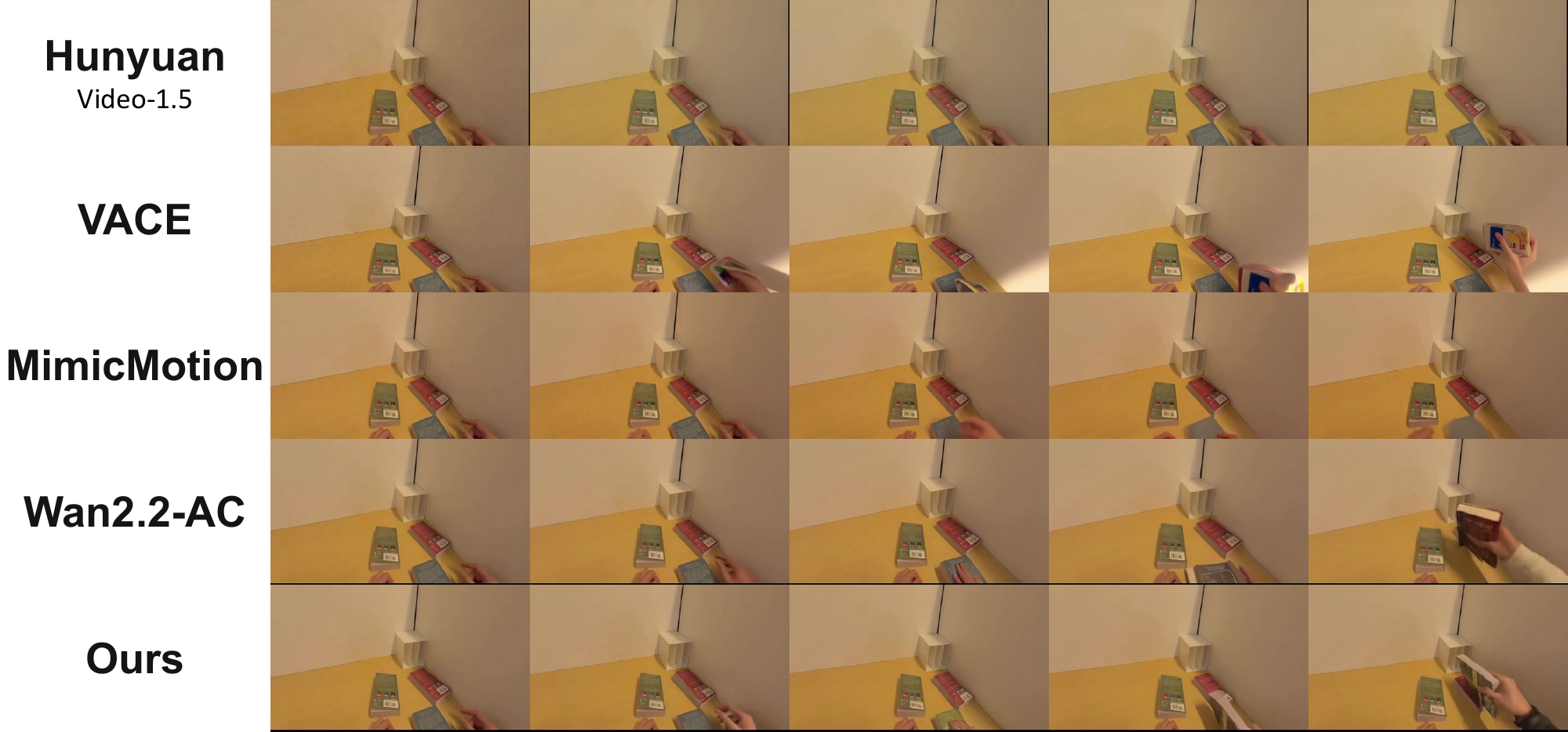}
\par\smallskip
{\normalsize\textbf{Prompt:} Pick up the \textcolor{red}{\textbf{red book}} from the
table.\par}
\caption{Book-pickup comparison. Rows show HunyuanVideo-1.5, VACE,
MimicMotion, Wan~2.2-AC, and \method{} (Ours) under the same initial observation
and task instruction; columns are uniformly sampled frames from each rollout.}
\label{fig:hand_case_06}
\end{figure}

\begin{figure}[p]
\centering
\appendixcaseimg{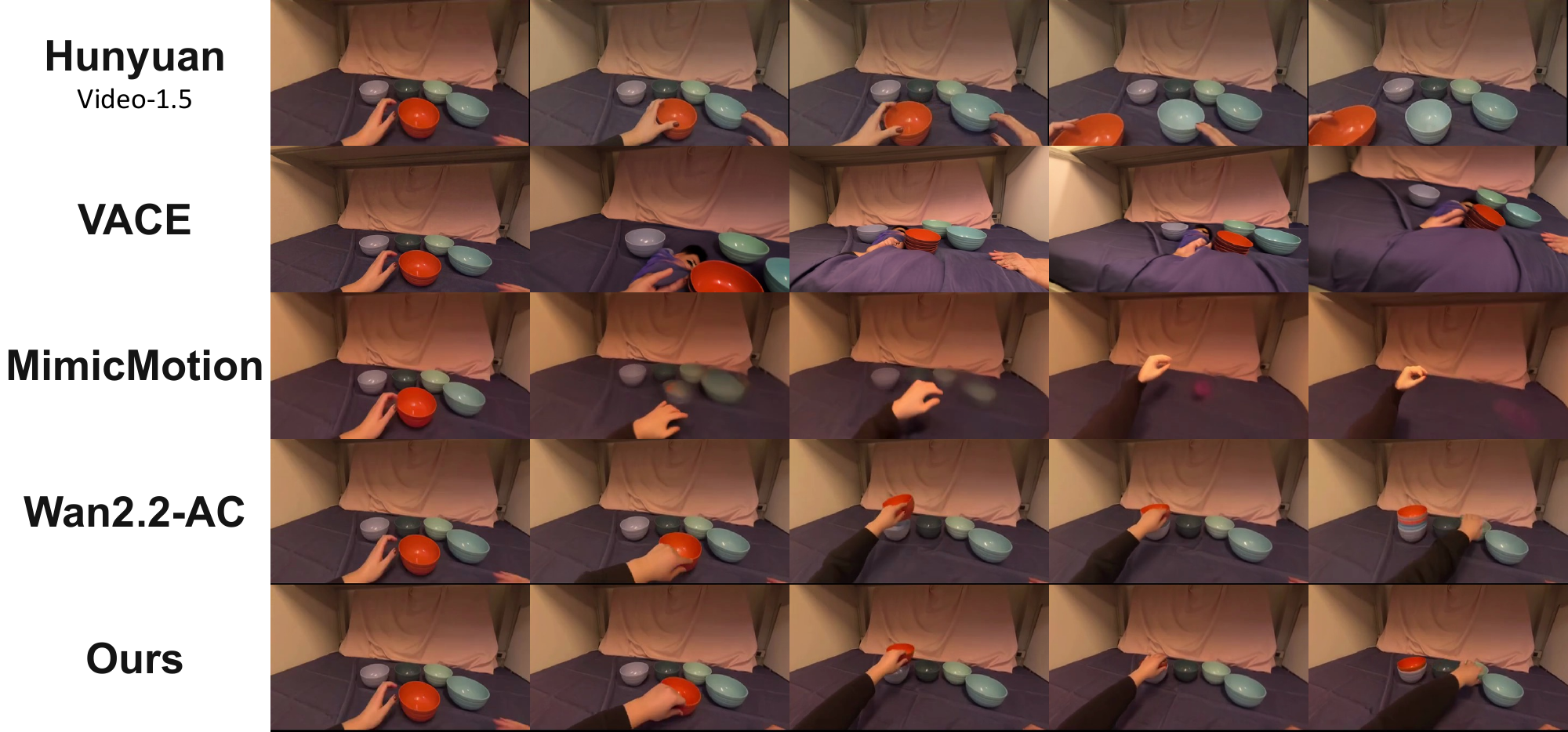}
\par\smallskip
{\normalsize\textbf{Prompt:} Stack the \textcolor{red}{\textbf{orange bowl}} on the
white bowl.\par}
\caption{Bowl-stacking comparison. Rows show HunyuanVideo-1.5, VACE,
MimicMotion, Wan~2.2-AC, and \method{} (Ours) under the same initial observation
and task instruction; columns are uniformly sampled frames from each rollout.}
\label{fig:hand_case_07}
\end{figure}

\end{document}